%% file: iclr2027_conference.tex
\documentclass{article} % For LaTeX2e
\usepackage{iclr2027_conference,times}

\input{math_commands.tex}

\usepackage{hyperref}
\usepackage{url}
\usepackage{enumitem}
\usepackage{graphicx}
\usepackage{booktabs}
\usepackage{wrapfig}
\usepackage{array}
\usepackage{tabularx}

\usepackage[table]{xcolor}
\definecolor{OpalStudentPurple}{RGB}{128,90,213}
\definecolor{OpalTeacherGreen}{RGB}{76,175,80}
\colorlet{OpalStudentBg}{OpalStudentPurple!10}
\colorlet{OpalTeacherBg}{OpalTeacherGreen!10}
\newcommand{\studentrow}{\rowcolor{OpalStudentBg}}
\newcommand{\teacherrow}{\rowcolor{OpalTeacherBg}}

\DeclareRobustCommand{\StudentTag}{{\setlength{\fboxsep}{1pt}\colorbox{OpalStudentBg}{\textcolor{black}{Student Model}}}}
\DeclareRobustCommand{\TeacherTag}{{\setlength{\fboxsep}{1pt}\colorbox{OpalTeacherBg}{\textcolor{black}{\textit{Teacher Model}}}}}

\usepackage{tcolorbox}
\tcbuselibrary{skins,breakable}

\usepackage{fvextra} 
\fvset{breaklines=true,breakanywhere=true}

\newcommand{\AlgName}{\textsc{Opal}}

\title{On-Policy Attention Linearization}

\author{%
\textbf{Arian Raje}\textsuperscript{1,\textdagger},
\textbf{Anupam Nayak}\textsuperscript{1,\textdagger},
\textbf{Anthony Fei}\textsuperscript{2}\\
\textbf{Akaash Parthasarathy}\textsuperscript{1},
\textbf{Mohamed Abdelfattah}\textsuperscript{2},
\textbf{Gauri Joshi}\textsuperscript{1}\\
\textsuperscript{1}Carnegie Mellon University\qquad
\textsuperscript{2}Cornell University\\
\textsuperscript{\textdagger}Equal contribution
}

\iclrfinalcopy 
\begin{document}

\maketitle

\input{Sections/Abstract}
\input{Sections/Introduction}
\input{Sections/Problem_Setup}
\input{Sections/Methods}
\input{Sections/Results}

\input{Sections/Conclusion}
\input{Sections/Post_Paper}
\bibliography{iclr2027_conference}
\bibliographystyle{iclr2027_conference}
\nocite{lan2026morphinghybridattentionmodels}

\newpage
\input{Sections/Appendix}

\end{document}

%% file: math_commands.tex
\usepackage{amsmath,amsfonts,bm}

\def\eqref#1{equation~\ref{#1}}
\def\1{\bm{1}}

\DeclareMathAlphabet{\mathsfit}{\encodingdefault}{\sfdefault}{m}{sl}
\SetMathAlphabet{\mathsfit}{bold}{\encodingdefault}{\sfdefault}{bx}{n}

%% file: Sections/Abstract.tex
\begin{abstract}
Hybrid transformer architectures that replace most softmax attention layers with linear attention offer transformer-level quality at a fraction of the memory cost. Rather than pretraining such models, a growing body of work distills them from already trained full-attention transformers. However, these distilled models often collapse on long-context retrieval and reasoning tasks, particularly when operating in thinking mode, where the efficiency gains of hybrid architectures matter most. Since linear attention layers must compress context into a fixed-size state, their errors compound over long sequences. As off-policy distillation never teaches the student model to recover from this drift, tasks that necessitate longer sequence lengths become especially challenging. We introduce \textbf{O}n-\textbf{P}olicy \textbf{A}ttention \textbf{L}inearization (\AlgName) in which the hybrid attention student samples its own long-context trajectories and receives dense supervision from the frozen full-attention teacher. Applying \AlgName\ to Qwen3-4B and MiMo-7B-RL-0530, we recover $87$--$94\%$ of full-attention performance on commonsense reasoning, $100\%$ on needle-in-a-haystack (NIAH) retrieval, and $83$--$93\%$ on mathematical reasoning with only 3B training tokens. We achieve these results without supervised fine-tuning (SFT) or reinforcement learning with verifiable rewards (RLVR). Compared with the strongest prior linearization method, which recovers $68\%$ of its teacher's retrieval performance and $21.6\%$ absolute average mathematical reasoning accuracy, \AlgName\ fully recovers retrieval and achieves $67.6$--$72.2\%$ on math reasoning.
\end{abstract}

%% file: Sections/Introduction.tex
\section{Introduction}
Attention-based large language models (LLMs) have become the de facto architecture for complex reasoning tasks \citep{NIPS2017_3f5ee243, openai2026openaio1card, guo2025deepseekr1}. A primary drawback of this architecture is that the traditional softmax-based implementation of attention, often referred to as full attention, is quadratic in complexity as each token ``attends'' to all prior tokens in the sequence. For long-context reasoning tasks, this can be prohibitively expensive. The KV cache, which maintains the key and value projections of every token, grows linearly with the number of tokens, ultimately dominating overall memory costs \citep{pope2023efficiently, kwon2023efficient}. To address the quadratic complexity of full attention, prior literature has proposed linear attention variants that maintain a constant-sized recurrent state rather than a growing KV cache, making per-token compute and memory independent of sequence length \citep{pmlr-v119-katharopoulos20a, gu2024mamba, sun2023retentivenetworksuccessortransformer, yang2024gated, yang2025gated}. Recently, open-source models have begun to incorporate hybrid attention where a small fraction of the transformer blocks maintain full-attention layers, but a majority of the transformer blocks utilize a linear attention variant, such as Mamba or Gated DeltaNet (GDN) \citep{ren2025samba, lenz2025jamba, qwenteam2026qwen35omnitechnicalreport, kimiteam2026kimik3openfrontier}. These hybrid models aim to balance the efficiency gains of linear attention with the long-context recall abilities of full attention. 

Recent work aims to distill linear attention or hybrid attention models from full-attention teachers, making it possible to create deployment-friendly versions of full-attention transformers without the exorbitant costs associated with pretraining \citep{zhang2025lolcats, goldstein2025radlads, bick2025llamba, yang2025zebrallama, chen2026hybridlinearattentionright, lan2026morphinghybridattentionmodels, li2026distilling}. Such methods generally proceed in two stages. First, the student's linear attention layers are initialized from the teacher's full-attention weights and trained to reproduce its attention hidden states. Second, the whole student model is fine-tuned to match the teacher's next-token distribution, typically using a KL objective on pretraining-like data. Prior to these two stages, an initial stage is sometimes included to determine which layers are most suitable for linear attention conversion \citep{zhou2026taylorcalibrateprincipledinitializationhybrid, bick2026retrievalaware}. 

However, while prior works benchmark their distilled linear attention or hybrid attention variants on commonsense/likelihood suites, the efficiency advantage of linear and hybrid attention only materializes when processing long inputs or generating many tokens, where using the recurrent state eliminates KV cache memory pressure. Thus, the settings that motivate linearization are long-context retrieval and multi-step reasoning, in which models must locate and process information across thousands of tokens. On precisely these tasks, we find that existing distilled models fall far short of their full attention teachers, even when they perform well on likelihood benchmarks.

\begin{figure*}[t]
  \centering
  \includegraphics[width=0.90\textwidth]{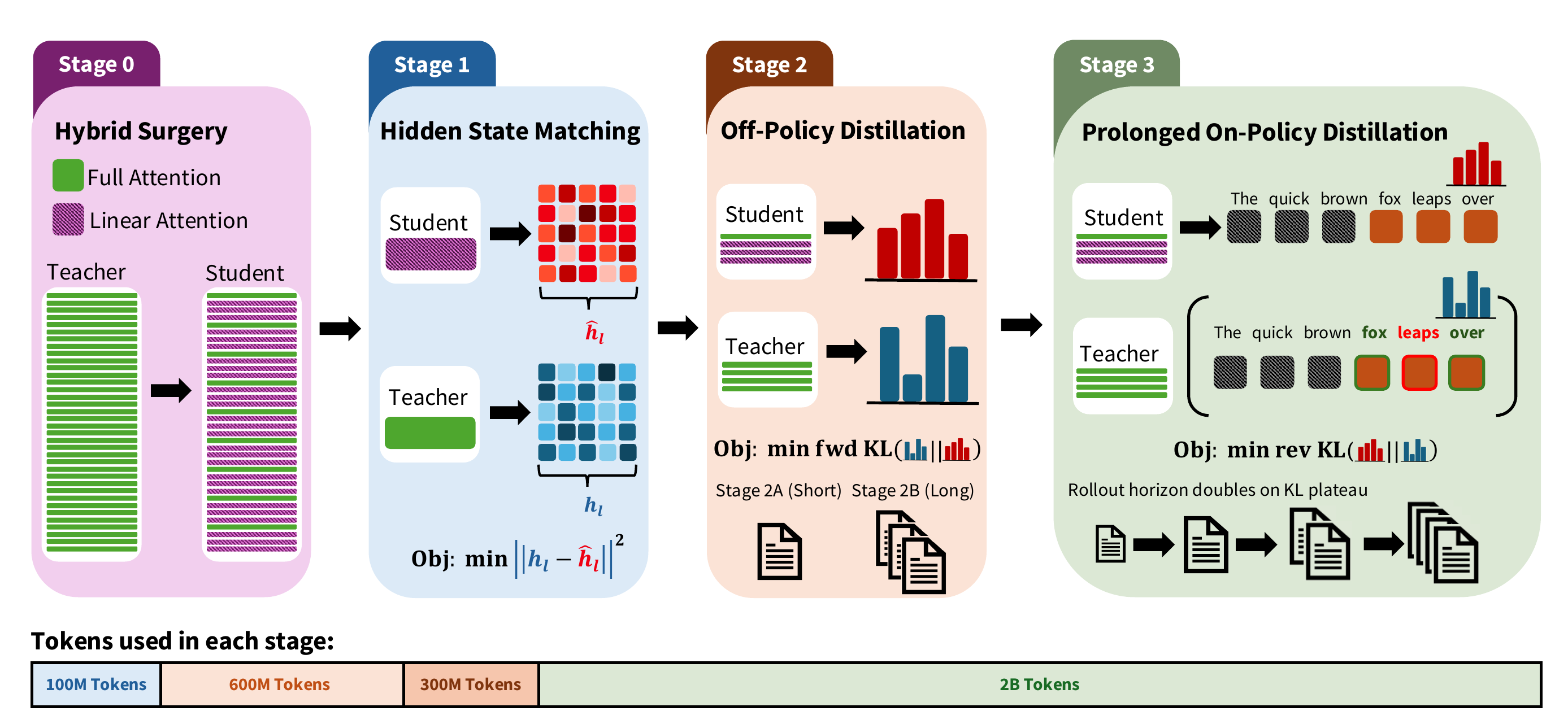}
  \caption{\textbf{Overview of \AlgName.} We convert a full attention teacher into a hybrid student by replacing $75\%$ of attention layers with linear attention. We recover the teacher's capabilities through hidden-state alignment, short- and long-context off-policy distillation, and prolonged on-policy distillation.}
  \label{fig:banner_opal}
  \vspace{-6mm}
\end{figure*}

We introduce \textbf{O}n-\textbf{P}olicy \textbf{A}ttention \textbf{L}inearization (\AlgName) a method for converting pretrained full-attention models into efficient hybrid attention models while preserving their long-context capabilities. \AlgName\ retains the standard two-stage recipe of prior work and appends a third stage of on-policy distillation (OPD) \citep{agarwal2024onpolicy, lu2025onpolicydistillation} in which the student is trained on sequences it generates rather than teacher-forced contexts. \AlgName\ proceeds in three key stages. First, we initialize the hybrid student's linear attention layers from the teacher's full-attention weights and align their hidden states. Next, we distill the hybrid student against the teacher's next-token distribution on a fixed corpus. Finally, the student samples its own continuations, the teacher scores every token, and the student minimizes the divergence from the teacher along its own trajectories (Figure ~\ref{fig:banner_opal}). We improve the efficacy of the third stage of this pipeline, the OPD stage, in three ways.

\begin{itemize}[leftmargin=*,itemsep=0pt,topsep=0pt]
    \item Training begins with short student rollouts and doubles their length each time the teacher-student KL plateaus. In this way, the student progressively improves on longer horizons. 
    \item We run OPD as a prolonged continual-learning-style scaling ladder using warmup-stable-decay (WSD) for the learning rate schedule \citep{hu2024minicpm, wen2025understanding}. This lets us extend training from 200M to 2B OPD tokens without restarting and yields checkpoints for measuring how reasoning recovers with on-policy compute. 
    \item We apply a difficulty curriculum over the reasoning prompts used for rollouts, introducing harder tasks as the student's accuracy on easier ones saturates over the course of the OPD checkpoints. 
\end{itemize}
We apply \AlgName\ to Qwen3-4B and MiMo-7B-RL-0530 and distill each into a hybrid student with $75\%$ of its full-attention layers replaced by GDN. With 2B OPD tokens, both students match their full-attention teachers on retrieval at every context length up to 32K, raise MATH-500 with thinking to $92\%$ and $93.8\%$ against teacher scores of $96.4\%$ and $97\%$, and recover $92.3\%$ and $100\%$ of teacher pass@8 on AIME'24. Across six prior distillation methods, the best baseline retains $68\%$ of its full-attention teacher's retrieval accuracy and $66.4\%$ of its reasoning average, while \AlgName\ students retain $100\%$ of retrieval and $83\%$ and $93\%$ reasoning. In absolute terms, \AlgName\ achieves a $67.6$--$72.2\%$ reasoning average, over $3\times$ the $21.6\%$ of the strongest prior linearized model. Our results show that a trained transformer's long-context retrieval and reasoning abilities can survive a late architectural conversion and that performance can be cheaply and efficiently recovered. 

%% file: Sections/Problem_Setup.tex
\section{Problem Setup and Motivation}
\textbf{Attention Architectures and Hybrid Transformers.} Given queries, keys, and values $q_t, k_t, v_t \in \mathbb{R}^d$ at token $t$ and with per-head attention dimension $d$, a softmax attention layer computes
\begin{equation}
o_t \;=\; \sum_{i \le t} \frac{\exp\!\left(q_t^\top k_i / \sqrt{d}\right)}{\sum_{j \le t} \exp\!\left(q_t^\top k_j / \sqrt{d}\right)}\, v_i .
\label{eq:softmax-attn}
\end{equation}
To compute softmax attention, the layer must maintain a history of all previous keys and values in a KV cache. Consequently, the KV cache and per-token compute grow linearly with sequence length. In contrast, linear attention replaces the exponential kernel with a feature map $\phi:\mathbb{R}^{d}\rightarrow\mathbb{R}^{d_\phi}$, so that $\exp(q^\top k/\sqrt{d}) \approx \phi(q)^\top \phi(k)$. Dropping the normalization from Equation \ref{eq:softmax-attn}, we get the following 
\begin{align}
o_t \;&=\; \sum_{i \le t} \left(\phi(q_t)^\top \phi(k_i)\right) v_i
\;=\; \sum_{i \le t} v_i\, \phi(k_i)^\top \phi(q_t)
\;=\; \underbrace{\left(\sum_{i \le t} v_i\, \phi(k_i)^\top\right)}_{S_t} \phi(q_t).
\label{eq:linear-factor}
\end{align}
Therefore, linear attention compresses the prefix into a recurrent state $S_t\in\mathbb{R}^{d\times d_\phi}$ so its per-token compute and state memory
remain $O(1)$ with respect to sequence length. Each token attends to the prefix through $S_t$ rather than through the exact keys and values. Linear attention variants differ in how they write to this state. Equation \ref{eq:linear-factor} implies an additive update $S_t = S_{t-1} + v_t \phi(k_t)^\top$, whereas the more expressive GDN layer updates $S_t$ according to the delta rule \citep{yang2024gated} as follows
\begin{equation}
S_t \;=\; \alpha_t\, S_{t-1}\left(I - \beta_t\, k_t k_t^\top\right) + \beta_t\, v_t k_t^\top,
\label{eq:gdn}
\end{equation}
where $\alpha_t \in (0,1)$ is a data-dependent decay term and $\beta_t \in (0,1)$ controls the update strength. For simplicity, we consider the common case $d_\phi=d$ and absorb the transformation into the notation $q_t$ and $k_t$. The combination gives the layer both global forgetting and targeted overwriting, allowing GDN to perform comparably to full attention. Hybrid attention architectures interleave a small number of softmax attention layers among linear attention layers. Some open-weight models, such as the Qwen3.5 family \citep{qwen3.5}, Kimi K3 \citep{kimiteam2026kimik3openfrontier}, and Nemotron 3 Nano \cite{nvidia2025nemotron3nanoopen} have begun adopting this model design to better balance efficiency and performance. 

\textbf{Distillation for Linearization.} 
Rather than pretraining linear or hybrid models from scratch, many works distill \citep{hinton2015distillingknowledgeneuralnetwork, kim2016sequencelevelknowledgedistillation} an existing full-attention transformer into a linear or hybrid student \citep{zhang2025lolcats, goldstein2025radlads, li2026distilling, lan2026morphinghybridattentionmodels}. These methods produce efficient variants of trained models at a small fraction of their original training cost. In practice, these methods follow a two-stage distillation recipe; they first align the outputs of each converted linear attention layer to the full-attention layer it replaces and then fine-tune the student against the teacher's next-token distribution. Let $p_T$ denote the teacher distribution, $p_\theta$ the student distribution, and $\mathcal{D}$ a fixed text corpus. The forward KL loss is given by
\begin{equation}
\mathcal{L}(\theta)
\;=\;
\mathbb{E}_{x \sim \mathcal{D}}
\sum_{t}
D_{\mathrm{KL}}\!\left(
p_T(\cdot \mid x_{<t})
\,\middle\|\,
p_\theta(\cdot \mid x_{<t})
\right),
\;
\text{where }D_{\mathrm{KL}}\!\left(p_T \,\middle\|\, p_\theta\right)
\;=\;
\mathbb{E}_{y \sim p_T}\!\left[\log \frac{p_T(y)}{p_\theta(y)}\right].
\label{eq:offpolicy}
\end{equation}
We call this objective \emph{off-policy} because the prefixes $x_{<t}$ on which the student is supervised are drawn from $\mathcal{D}$ and never from the student itself. Note that $\mathcal{D}$ may consist of teacher generations, which is still off-policy since the prefixes are also not the student's own traces. Ultimately, in either case, Equation \ref{eq:offpolicy} provides no direct
supervision on prefixes sampled from the student's own policy. Prior linearization methods, such as RADLADS \citep{goldstein2025radlads}, Zebra-Llama \citep{yang2025zebrallama}, and HALO \citep{chen2026hybridlinearattentionright}, all optimize Equation \ref{eq:offpolicy} or a close variant on a fixed corpus, even if they differ in their initialization of converted attention layers or the way they schedule the conversion in their pipelines.

\begin{figure}[t]
  \centering
  % Locally reduce caption-to-text spacing for this figure only.
  \includegraphics[width=\linewidth]{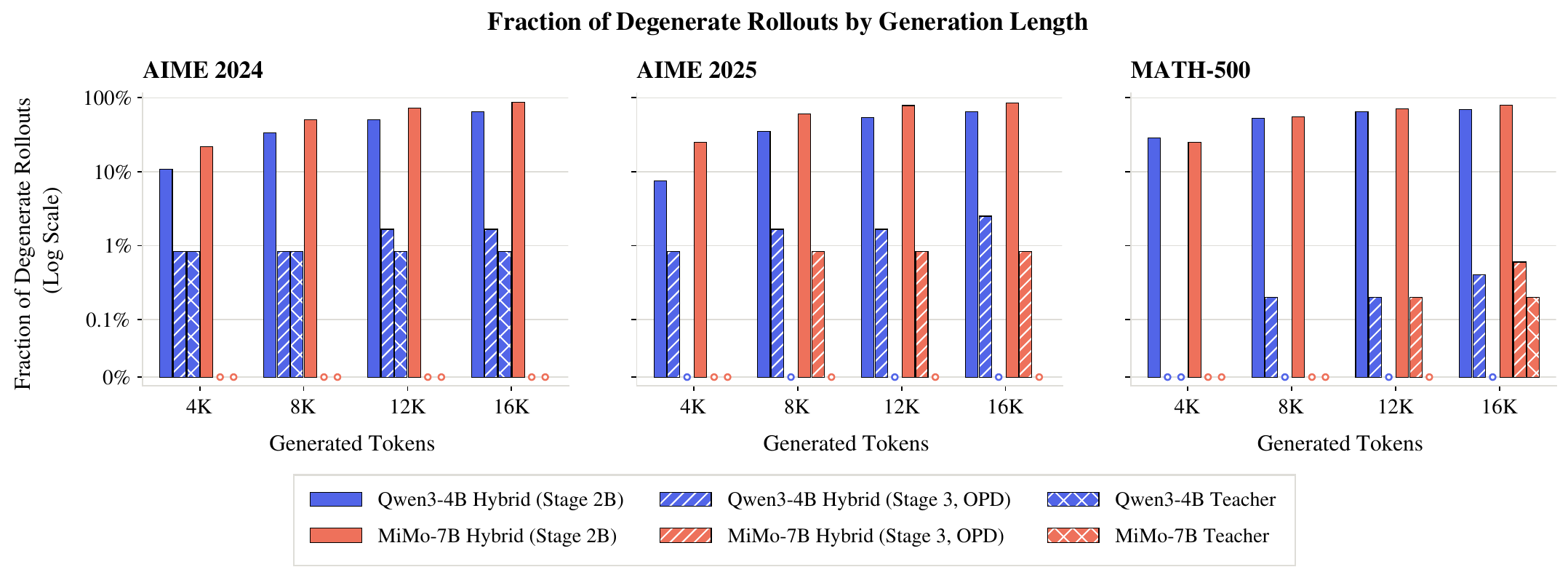}
  \caption{Degenerate rollouts during student decoding. Off-policy distillation aligns the student on fixed prefixes, but provides no signal on the student's own trajectories. Student rollouts can collapse into degenerate rollouts such as repetitive text or unrelated prose. Open circles indicate $0\%$.}
  \vspace{-4mm}
  \label{fig:degenerate-rollouts}
\end{figure}

\textbf{Distilled Hybrid Students on Long-Context Tasks.}
Consider a transformer with $L$ layers and per-head dimension $d$ applied to a length $n$ sequence. For a full-attention model, cache memory is
\begin{equation} 
M_{\mathrm{FA}}(n) = O(Lnd), \label{eq:full-attn-memory} \end{equation}
omitting attention heads. Now consider a hybrid model in which only $L_{\mathrm{FA}}$ layers retain full attention. Since linear attention layers store a fixed-size recurrent state $S_t$, its memory scales as \begin{equation} M_{\mathrm{Hybrid}}(n) = O(L_{\mathrm{FA}}nd) + O((L-L_{\mathrm{FA}})d^2). \label{eq:hybrid-memory} \end{equation}
The first term grows with context length, while the second remains fixed. As $n$ increases, the memory savings from replacing full-attention layers become increasingly significant. 

However, strong short-context performance after linearization does not guarantee long-context robustness. As shown in Figure \ref{fig:degenerate-rollouts}, distilled hybrid students that are trained with attention matching and off-policy distillation can diverge from their teachers over long generations and fall into repetitive or degenerate trajectories. For linearized models, once an error changes the generated prefix, it also changes the recurrent state $S_t$ that compresses that history. This ultimately affects every subsequent prediction. Because off-policy distillation in Equation \ref{eq:offpolicy} only supervises fixed prefixes from $\mathcal{D}$, the student is never explicitly trained to recover from states induced by its own mistakes. Over long horizons, these deviations can compound. Our goal is to linearize trained transformers without sacrificing the long-context retrieval and reasoning abilities that make efficient decoding valuable in the first place. This requires training the student to remain well-behaved on its own trajectories.

%% file: Sections/Methods.tex
\section{Proposed Method}
\AlgName\ converts a trained full-attention model into an efficient hybrid model while preserving its learned representations and behavior. Starting from the full-attention teacher, we construct a student parameterized by $\theta$ by retaining every fourth full-attention layer and replacing the remaining layers with GDN layers. For the 36-layer models considered in this work, the resulting hybrid architecture contains 9 interleaved full attention layers and 27 GDN layers (Figure \ref{fig:banner_opal}). All non-attention components, such as the embeddings, MLPs, normalization layers, and output head, are inherited directly from the teacher. Compatible full-attention projections are used to initialize the corresponding GDN projections. The converted layers also introduce the GDN-specific forget and update gates $\alpha$ and $\beta$, an output gate, and a short causal convolution \citep{yang2024gated}. 

Architectural conversion substantially affects the model's representations and next-token distribution. To address this issue, we recover the teacher through the three training stages illustrated in Figure~\ref{fig:banner_opal}. We align the representations produced by the converted layers, optimize $\theta$ to match the teacher on fixed short- and long-context sequences, and train on student sampled trajectories.

\textbf{Stage 1: Hidden State Matching.} We first align each converted GDN layer independently with its corresponding teacher layer. For a sequence $\mathbf{x}=(x_1,\ldots,x_T)\sim\mathcal{D}$, let $h_{\ell,t}$ and
$\hat{h}_{\ell,t}$ denote the teacher and student hidden states at layer
$\ell$ and position $t$. We optimize the following 
\begin{equation}
    \mathcal{L}_{\mathrm{match}}(\theta)
    =
    \frac{1}{|L_{\mathrm{GDN}}|}
    \sum_{\ell \in L_{\mathrm{GDN}}}
    \mathbb{E}_{\mathbf{x}\sim\mathcal{D}}
    \left[
        \frac{1}{T}
        \sum_{t=1}^{T}
        \left\|
            h_{\ell,t} - \hat{h}_{\ell,t}
        \right\|_2^2
    \right].
    \label{eq:hidden_alignment}
\end{equation}
This loss isolates the architectural mismatch introduced by replacing full attention with GDN. The matching stage ensures the model outputs coherent, albeit inaccurate and unrelated, text.
During this stage, only the parameters of the converted GDN layers are updated. The retained full-attention layers and all other model parameters are frozen (equivalently, assigned 0 learning rate).

\textbf{Stage 2: Off-Policy Distillation.} Hidden state matching trains each converted layer independently but does not ensure that the entire hybrid model reproduces the teacher's next-token distribution. In Stage 2, we train all of the hybrid student's model weights with a short off-policy distillation stage. We use the objective in Equation~\ref{eq:offpolicy}, matching the student's predictive distribution to the teacher's at each position along ground-truth corpus sequences.  This stage aims to remedy errors introduced by composing the individually aligned GDN layers. We first distill on short sequences 4K in length (Stage 2A) before extending to longer sequences 32K in length (Stage 2B) as a curriculum for improving on longer contexts. We train the entire hybrid model with two different step sizes, using a learning rate for retained full-attention layers that is $10\times$ smaller than for GDN layers. 

Despite training on different-length sequences during Stage 2, we view this stage primarily as a warm start for OPD rather than the mechanism responsible for long-context recovery. Because every training prefix remains drawn from a fixed corpus, the student is never exposed to states induced by its own predictions. Therefore, off-policy distillation can provide a well-calibrated starting point, but it does not directly address the compounding state drift that arises during long rollouts.

\textbf{Stage 3: Prolonged On-Policy Distillation.} While prior works have considered analogues of Stages 1 and 2, our novel addition to the linearization pipeline is a stage of prolonged OPD. As mentioned, during generation, the student must condition on its own previous predictions. Thus, prolonged OPD is the only stage aimed at addressing the mismatch between prior objectives and model inference. Given a prompt $\mathbf{x}$, the student samples a continuation $\mathbf{y}=(y_1,\ldots,y_H)$ from its current policy. The frozen teacher is then applied to this trajectory, and we minimize the token-level reverse KL 
\begin{equation}
    \mathcal{L}_{\mathrm{OPD}}(\theta)
    =
    \mathbb{E}_{\mathbf{x}\sim\mathcal{D},
    \,\mathbf{y}\sim p_\theta(\cdot\mid\mathbf{x})}
    \left[
        \frac{1}{H}
        \sum_{t=1}^{H}
        D_{\mathrm{KL}}
        \left(
            p_\theta(\cdot\mid\mathbf{x},\mathbf{y}_{<t})
            \,\Vert\,
            p_T(\cdot\mid\mathbf{x},\mathbf{y}_{<t})
        \right)
    \right].
    \label{eq:opd}
\end{equation}
The loss is applied only to student-generated positions, meaning the prompt does not contribute to the loss. With this loss, the student recurrent states $S_t$ that are generated during decoding are no longer outside the training distribution as the teacher supervises the student after its own deviations. 

\emph{KL-Guided Rollout Horizon Doubling.}
Long rollouts are expensive and especially unstable early in OPD when the student already diverges from the teacher over short horizons. Instead, we adaptively increase rollout length throughout training. Training begins at $H=512$,
and doubles as follows,
\begin{equation}
    \Delta D_{\mathrm{KL}}\!\left(p_\theta \,\|\, p_T\right)
    < \epsilon \;\text{ for } P \text{ updates}
    \quad\Longrightarrow\quad
    H \leftarrow \min(2H,H_{\max}),
    \label{eq:horizon_schedule}
\end{equation}
where $H_{\max} =$ 32K and $\Delta D_{\mathrm{KL}}$ denotes the relative improvement of the reverse KL over its best value at the current rollout horizon. By increasing the rollout horizon only after KL plateaus, we train the student to remain stable over progressively longer sequences of recurrent state updates.

\emph{Sample Curriculum.}
We also progressively increase the difficulty of the OPD training distribution. Note that difficulty is controlled at the dataset level rather than estimated per sample. Once the rollout horizon reaches $1$K, we introduce long-context retrieval examples spanning context lengths of $2$K-$32$K tokens alongside reasoning and general instruction-following prompts. At later OPD checkpoints, we further replace the initial reasoning data with harder mathematical problems. 

\emph{Prolonged OPD with WSD.}
Long-horizon capabilities continue to improve as we increase the amount of on-policy training. But, the required training budget is difficult to determine a priori. To enable effective checkpointing, we use a WSD learning rate schedule that lets us incrementally extend OPD while retaining annealed checkpoints at intermediate budgets. At each target token budget $B_k$, we save the full training state immediately before the decay phase. In this way, we can branch from this predecayed checkpoint to both evaluate the checkpoint and continue OPD training 
\begin{equation}
\theta_k^{\mathrm{pre}}
\;\begin{cases}
\xrightarrow{\;\mathrm{decay}\;} \theta_k^{\mathrm{eval}},\\[3pt]
\xrightarrow{\;\mathrm{extend}\;} \theta_{k+1}^{\mathrm{pre}}
\xrightarrow{\;\mathrm{decay}\;} \theta_{k+1}^{\mathrm{eval}},
\end{cases}
\qquad B_{k+1} > B_k .
\label{eq:wsd_branch}
\end{equation}
This procedure lets us scale a single OPD trajectory to 2B OPD tokens while obtaining comparable checkpoints throughout. Beyond extending the Stage 3 training formula efficiently, it allows for direct measurement of long-context retrieval and reasoning recover as a function of OPD compute.

Unlike standard OPD settings that begin from an already long-context capable student, our post-Stage 2 student remains substantially weaker than the teacher on long contexts. Consequently, early student rollouts frequently depart from regions where the teacher and student behave similarly, and recovery requires sustained on-policy training as the student's trajectory distribution evolves. We use the original full-attention model as the teacher rather than a stronger model as our objective is to recover capabilities lost during architectural conversion rather than train capabilities that the original model did not possess. This combination of an initially weak student and the use of the original full-attention model, rather than a stronger teacher, makes prolonging the OPD stage necessary.

As the only layers with direct access to the full token history, the retained full attention layers must support long-range interaction in ways that the GDN layers with their compressed state cannot. While prior approaches often keep these retained attention layers frozen during distillation, we keep them trainable. Stage 2 uses two step sizes, before Stage 3 switches to the same flat learning rate for all layers so that the GDN and retained full-attention layers can train jointly.

%% file: Sections/Results.tex
\section{Results}
\subsection{Experimental Setup}
\textbf{Models and Datasets.} We apply \AlgName\ to Qwen3-4B and MiMo-7B-RL-0530. Both contain 36 transformer layers, so our linearization pipeline results in 9 full attention layers and 27 GDN layers. Stages 1 and 2 use a pretraining-style mixture of FineWeb-Edu \citep{penedo2024the}, CodeParrot-Clean, and OpenWebMath \citep{paster2024openwebmath}. For OPD, we sample prompts from OpenThoughts-114K-Math \citep{guha2026openthoughts}, Dolly-15K \citep{DatabricksBlog2023DollyV2}, and synthetic long-context retrieval tasks. Subsequent checkpoints are also trained with DeepScaleR \citep{deepscaler2025} and DAPO-Math-17K \citep{yu2025dapo} in place of the OpenThoughts-114K-Math to encourage performance on harder tasks. The datasets are filtered by 13-gram overlap with our evaluation tasks. 

We evaluate three aspects of model quality. For short-context retention, we use PIQA \citep{bisk2020piqa}, HellaSwag \citep{zellers-etal-2019-hellaswag}, ARC-Easy \citep{clark2018thinksolvedquestionanswering}, ARC-Challenge, WinoGrande \citep{10.1145/3474381}, and MMLU \citep{hendrycks2021measuring}, using zero-shot evaluation except for 5-shot MMLU. For long-context retrieval, we evaluate single needle, multi-key, and multi-query RULER-style tasks \citep{hsieh2024ruler} at 4K, 8K, 16K, and 32K context lengths. Finally, we evaluate mathematical reasoning on GSM8K \citep{cobbe2021trainingverifierssolvemath}, GSM-Plus \citep{li2024gsmplus}, GSM-Symbolic \citep{mirzadeh2025gsmsymbolic}, MATH-500, AIME'24 \citep{aime24}, and AIME'25 \citep{aime25}. We report pass@1 for GSM8K, GSM-Plus, GSM-Symbolic, and MATH-500 and pass@1 and pass@8 for AIME, the most challenging reasoning task.

\textbf{Training Parameters.} Each student is trained on approximately 3B tokens: 100M for attention alignment, 600M for short-context distillation, 300M for 32K long-context distillation, and 2B student-generated tokens for OPD. Alignment and short-context KD use 4K sequences, while long-context KD uses 32K sequences. During OPD, rollouts begin at 512 tokens and iteratively increase in length until 32K. Complete optimization hyperparameters are provided in the Appendix. 

\textbf{Baselines.} We compare against publicly released linearized models from RADLADS \citep{goldstein2025radlads}, Llamba \citep{bick2025llamba}, Zebra-Llama \citep{yang2025zebrallama}, and HALO \citep{chen2026hybridlinearattentionright}. We compare each student with the full-attention teacher from which it is distilled. Because these approaches differ substantially in teacher strength and training budget, we emphasize recovery relative to the full-attention teacher in addition to absolute performance. To isolate the effect of OPD, we also train off-policy controls using the same Qwen3-4B hybrid initialization and training schedule, but instead distill on fixed dataset sequences, similar to the Stage 2B portion.

\begin{table*}[t]
    \centering

    \renewcommand{\arraystretch}{0.92}

    \caption{Needle-in-a-Haystack retrieval accuracy (\%). \TeacherTag; \StudentTag.}
    \label{tab:niah}
    \setlength{\tabcolsep}{3.5pt}
    \resizebox{\textwidth}{!}{%
    \begin{tabular}{lccccccccccccc}
        \toprule
        & \multicolumn{4}{c}{\textbf{Single Needle}}
        & \multicolumn{4}{c}{\textbf{Multi-Key}}
        & \multicolumn{4}{c}{\textbf{Multi-Query}}
        & \\
        \cmidrule(lr){2-5}
        \cmidrule(lr){6-9}
        \cmidrule(lr){10-13}
        \textbf{Model}
        & \textbf{4K} & \textbf{8K} & \textbf{16K} & \textbf{32K}
        & \textbf{4K} & \textbf{8K} & \textbf{16K} & \textbf{32K}
        & \textbf{4K} & \textbf{8K} & \textbf{16K} & \textbf{32K}
        & \textbf{Avg.} \\
        \midrule

        \teacherrow
        \textit{Qwen2.5-7B-Instruct}
            & $100.0$ & $100.0$ & $100.0$ & $100.0$
            & $100.0$ & $100.0$ & $100.0$ & $100.0$
            & $100.0$ & $100.0$ & $100.0$ & $100.0$
            & $100.0$ \\
        \studentrow
        QRWKV6-7B-Instruct
            & $0.0$ & $0.0$ & $0.0$ & $0.0$
            & $1.0$ & $0.0$ & $0.0$ & $0.0$
            & $0.0$ & $0.0$ & $0.0$ & $0.0$
            & $0.1$ {\scriptsize ($0.1\%$)} \\

        \midrule
        \teacherrow
        \textit{Llama-3.2-1B-Instruct}
            & $100.0$ & $100.0$ & $100.0$ & $100.0$
            & $94.0$ & $93.0$ & $71.0$ & $67.0$
            & $98.5$ & $94.2$ & $93.2$ & $96.5$
            & $92.3$ \\
        \studentrow
        Llamba-1B
            & $0.0$ & $0.0$ & $0.0$ & $0.0$
            & $0.0$ & $0.0$ & $0.0$ & $0.0$
            & $0.0$ & $0.0$ & $0.0$ & $0.0$
            & $0.0$ {\scriptsize ($0.0\%$)} \\

        \midrule
        \teacherrow
        \textit{Llama-3.2-3B-Instruct}
            & $100.0$ & $100.0$ & $100.0$ & $100.0$
            & $98.0$ & $95.0$ & $94.0$ & $87.0$
            & $100.0$ & $99.2$ & $87.5$ & $68.5$
            & $94.1$ \\
        \studentrow
        Llamba-3B
            & $2.0$ & $0.0$ & $0.0$ & $0.0$
            & $3.0$ & $3.0$ & $0.0$ & $0.0$
            & $0.8$ & $0.0$ & $0.0$ & $0.0$
            & $0.7$ {\scriptsize ($0.8\%$)} \\
        \studentrow
        Zebra-Llama-3B
            & $85.0$ & $61.0$ & $52.0$ & $47.0$
            & $55.0$ & $39.0$ & $28.0$ & $10.0$
            & $64.2$ & $24.2$ & $23.0$ & $3.5$
            & $41.0$ {\scriptsize ($43.6\%$)} \\

        \midrule
        \teacherrow
        \textit{Llama-3.1-8B-Instruct}
            & $100.0$ & $100.0$ & $100.0$ & $100.0$
            & $100.0$ & $100.0$ & $99.0$ & $99.0$
            & $100.0$ & $100.0$ & $100.0$ & $100.0$
            & $99.8$ \\
        \studentrow
        Zebra-Llama-8B
            & $\mathbf{100.0}$ & $98.0$ & $96.0$ & $74.0$
            & $73.0$ & $40.0$ & $40.0$ & $27.0$
            & $86.8$ & $70.0$ & $66.5$ & $43.0$
            & $67.9$ {\scriptsize ($68.0\%$)} \\

        \midrule
        \teacherrow
        \textit{Qwen3-1.7B}
            & $100.0$ & $100.0$ & $100.0$ & $100.0$
            & $99.0$ & $100.0$ & $99.0$ & $99.0$
            & $100.0$ & $100.0$ & $100.0$ & $100.0$
            & $99.8$ \\
        \studentrow
        HypeNet-2B
            & $73.0$ & $76.0$ & $66.0$ & $73.0$
            & $28.0$ & $22.0$ & $24.0$ & $18.0$
            & $27.3$ & $19.0$ & $18.5$ & $15.5$
            & $38.4$ {\scriptsize ($38.5\%$)} \\

        \midrule
        \teacherrow
        \textit{Qwen3-4B}
            & $100.0$ & $100.0$ & $100.0$ & $100.0$
            & $100.0$ & $100.0$ & $100.0$ & $99.0$
            & $100.0$ & $100.0$ & $100.0$ & $100.0$
            & $99.9$ \\
        \studentrow
        HypeNet-5B
            & $88.0$ & $90.0$ & $90.0$ & $96.0$
            & $34.0$ & $37.0$ & $33.0$ & $28.0$
            & $49.5$ & $55.8$ & $49.2$ & $45.0$
            & $58.0$ {\scriptsize ($58.0\%$)} \\
        \studentrow
        OPAL-Qwen3 (ours)
            & $\mathbf{100.0}$ & $\mathbf{100.0}$ & $\mathbf{100.0}$ & $\mathbf{100.0}$
            & $\mathbf{100.0}$ & $\mathbf{100.0}$ & $\mathbf{100.0}$ & $\mathbf{99.6}$
            & $\mathbf{100.0}$ & $\mathbf{100.0}$ & $\mathbf{100.0}$ & $\mathbf{100.0}$
            & $\mathbf{100.0}$ {\scriptsize ($100.0\%$)} \\

        \midrule
        \teacherrow
        \textit{MiMo-7B-RL-0530}
            & $100.0$ & $100.0$ & $100.0$ & $100.0$
            & $99.6$ & $99.0$ & $98.8$ & $92.8$
            & $99.9$ & $99.7$ & $98.3$ & $76.2$
            & $97.0$ \\
        \studentrow
        OPAL-MiMo (ours)
            & $\mathbf{100.0}$ & $\mathbf{100.0}$ & $\mathbf{100.0}$ & $\mathbf{100.0}$
            & $96.8$ & $96.8$ & $94.8$ & $83.8$
            & $\mathbf{100.0}$ & $\mathbf{100.0}$ & $\mathbf{100.0}$ & $\mathbf{100.0}$
            & $97.7$ {\scriptsize ($100.7\%$)} \\

        \bottomrule
    \end{tabular}%
    }
    \renewcommand{\arraystretch}{1.0}
    \vspace{-6mm}
\end{table*}

\begin{table*}[t]
    \centering
    \caption{Mathematical reasoning performance (\%). \TeacherTag; \StudentTag.}
    \label{tab:math}
    \setlength{\tabcolsep}{6pt}
    \renewcommand{\arraystretch}{1.0}
    \resizebox{\textwidth}{!}{\scriptsize%
    \begin{tabular}{lccccccccc}
        \toprule
        & & & &
        & \multicolumn{2}{c}{\textbf{AIME'24}}
        & \multicolumn{2}{c}{\textbf{AIME'25}}
        & \\
        \cmidrule(lr){6-7}
        \cmidrule(lr){8-9}
        \textbf{Model}
        & \textbf{GSM8K}
        & \textbf{GSM-Plus}
        & \textbf{GSM-Sym}
        & \textbf{MATH-500}
        & \textbf{pass@1}
        & \textbf{pass@8}
        & \textbf{pass@1}
        & \textbf{pass@8}
        & \textbf{Avg.} \\
        \midrule

        \teacherrow
        \textit{Qwen2.5-7B-Instruct}
        & $91.9$ & $72.7$ & $83.4$ & $76.4$
        & $10.4$ & $26.7$ & $7.5$ & $23.3$
        & $49.0$ \\

        \studentrow
        QRWKV6-7B-Instruct
        & $27.4$ & $16.8$ & $9.3$ & $24.0$
        & $0.4$ & $3.3$ & $0.4$ & $3.3$
        & $10.6$ {\scriptsize ($21.6\%$)} \\

        \midrule

        \teacherrow
        \textit{Llama-3.2-1B-Instruct}
        & $38.5$ & $24.9$ & $26.2$ & $25.8$
        & $0.8$ & $6.7$ & $0.0$ & $0.0$
        & $15.4$ \\

        \studentrow
        Llamba-1B
        & $28.1$ & $12.9$ & $13.0$ & $8.6$
        & $0.0$ & $0.0$ & $0.0$ & $0.0$
        & $7.8$ {\scriptsize ($51.0\%$)} \\

        \midrule

        \teacherrow
        \textit{Llama-3.2-3B-Instruct}
        & $71.1$ & $58.3$ & $59.5$ & $39.0$
        & $2.5$ & $13.3$ & $0.0$ & $0.0$
        & $30.5$ \\

        \studentrow
        Llamba-3B
        & $50.7$ & $27.5$ & $24.8$ & $7.8$
        & $0.4$ & $3.3$ & $0.0$ & $0.0$
        & $14.3$ {\scriptsize ($47.0\%$)} \\

        \studentrow
        Zebra-Llama-3B
        & $62.7$ & $38.8$ & $36.7$ & $20.0$
        & $0.4$ & $3.3$ & $0.0$ & $0.0$
        & $20.2$ {\scriptsize ($66.4\%$)} \\

        \midrule

        \teacherrow
        \textit{Llama-3.1-8B-Instruct}
        & $86.1$ & $67.0$ & $73.2$ & $48.6$
        & $2.9$ & $10.0$ & $0.0$ & $0.0$
        & $36.0$ \\

        \studentrow
        Zebra-Llama-8B
        & $61.3$ & $40.3$ & $35.3$ & $24.4$
        & $0.8$ & $6.7$ & $0.4$ & $3.3$
        & $21.6$ {\scriptsize ($59.9\%$)} \\

        \midrule

        \teacherrow
        \textit{Qwen3-1.7B}
        & $83.5$ & $63.3$ & $65.0$ & $74.6$
        & $13.8$ & $30.0$ & $10.4$ & $23.3$
        & $45.5$ \\

        \studentrow
        HypeNet-2B
        & $1.1$ & $3.8$ & $1.2$ & $4.4$
        & $0.0$ & $0.0$ & $0.8$ & $3.3$
        & $1.8$ {\scriptsize ($4.0\%$)} \\

        \midrule

        \teacherrow
        \textit{Qwen3-4B}
        & $92.3$ & $72.8$ & $83.2$ & $85.4$
        & $25.8$ & $43.3$ & $22.1$ & $43.3$
        & $58.5$ \\

        \studentrow
        HypeNet-5B
        & $14.9$ & $15.8$ & $8.7$ & $11.8$
        & $0.0$ & $0.0$ & $0.0$ & $0.0$
        & $6.4$ {\scriptsize ($10.9\%$)} \\

        \midrule

        \teacherrow
        \textit{Qwen3-4B}
        & $92.3$ & $72.8$ & $83.2$ & $96.4$
        & $73.8$ & $86.7$ & $63.7$ & $80.0$
        & $81.1$ \\

        \studentrow
        OPAL-Qwen3 (ours)
        & $83.2$ & $\mathbf{63.8}$ & $\mathbf{65.9}$ & $92.0$
        & $52.5$ & $80.0$ & $40.4$ & $63.3$
        & $67.6$ {\scriptsize ($83.4\%$)} \\

        \midrule

        \teacherrow
        \textit{MiMo-7B-RL-0530}
        & $80.2$ & $61.6$ & $71.6$ & $97.0$
        & $73.3$ & $83.3$ & $67.9$ & $83.3$
        & $77.3$ \\

        \studentrow
        OPAL-MiMo (ours)
        & $\mathbf{84.8}$ & $63.5$ & $65.7$ & $\mathbf{93.8}$
        & $\mathbf{65.4}$ & $\mathbf{83.3}$ & $\mathbf{47.5}$ & $\mathbf{73.3}$
        & $\mathbf{72.2}$ {\scriptsize ($93.4\%$)} \\

        \bottomrule
    \end{tabular}%
    }
    \vspace{-4mm}
\end{table*}

\subsection{Main Results}
In Tables~\ref{tab:niah} and~\ref{tab:math}, we demonstrate the performance of our \AlgName-linearized models relative to prior linearization baselines for NIAH and math reasoning tasks. On NIAH, OPAL-Qwen3 matches its teacher with an average accuracy of $100.0\%$. The strongest baseline, Zebra-Llamba-8B only recovers $68\%$ of its teacher's performance, $32\%$ lower than both of our linearized models. These gains also extend to mathematical reasoning. OPAL-MiMo retains $93\%$ of its teacher's average math reasoning performance and completely recovers AIME'24 pass@8. Its absolute math reasoning average is also $50.6\%$ higher than the next strongest baseline. As HypeNet-2B and HypeNet-5B are not trained with thinking mode enabled, they must be compared to the performance of their non-thinking teachers. However, HypeNet-5B, which has the same teacher model as OPAL-Qwen3, lags behind OPAL-Qwen3's average math reasoning performance by $61.2\%$.  

This improved long-context performance comes with a tradeoff on commonsense reasoning benchmarks, whose full results we defer to the Appendix for space. OPAL-Qwen3 retains $87\%$ of its teacher's performance on the commonsense suite, while OPAL-MiMo retains $94\%$ with absolute scores remaining within the range achieved by prior linearized models. Although we acknowledge the tradeoff in commonsense performance, these results reflect a shift toward preserving the long-context capabilities that motivate hybrid attention in the first place. \AlgName\ achieves these improvements with a comparatively small training budget and without separate post-training phases beyond distillation. RADLADS and HALO both use SFT for post-training, whereas Zebra-Llama incorporates a short DPO stage after SFT. Zebra-Llama-3B and Zebra-Llama-8B use 9B and 11B training tokens, respectively. In contrast, OPAL-Qwen3 and OPAL-MiMo only use 3B training tokens.
\begin{figure}[t]
  \centering
  \includegraphics[width=0.99\linewidth]{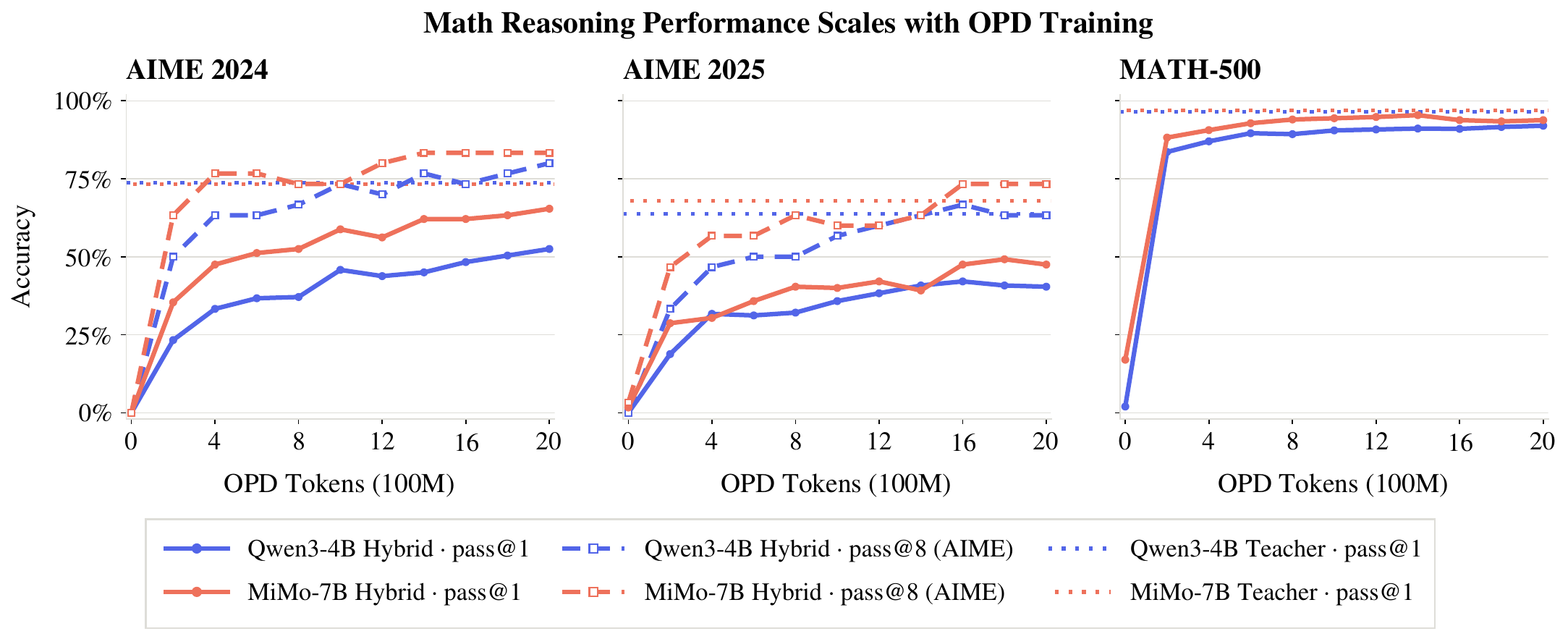}
  \caption{Math reasoning performance as a function of OPD training tokens. MATH-500 improves rapidly and saturates early, while AIME continues to benefit from additional OPD compute.}
  \label{fig:math_scaling}
  \vspace{-2mm}
\end{figure}
\begin{figure}[t]
  \centering
  \includegraphics[width=0.99\linewidth]{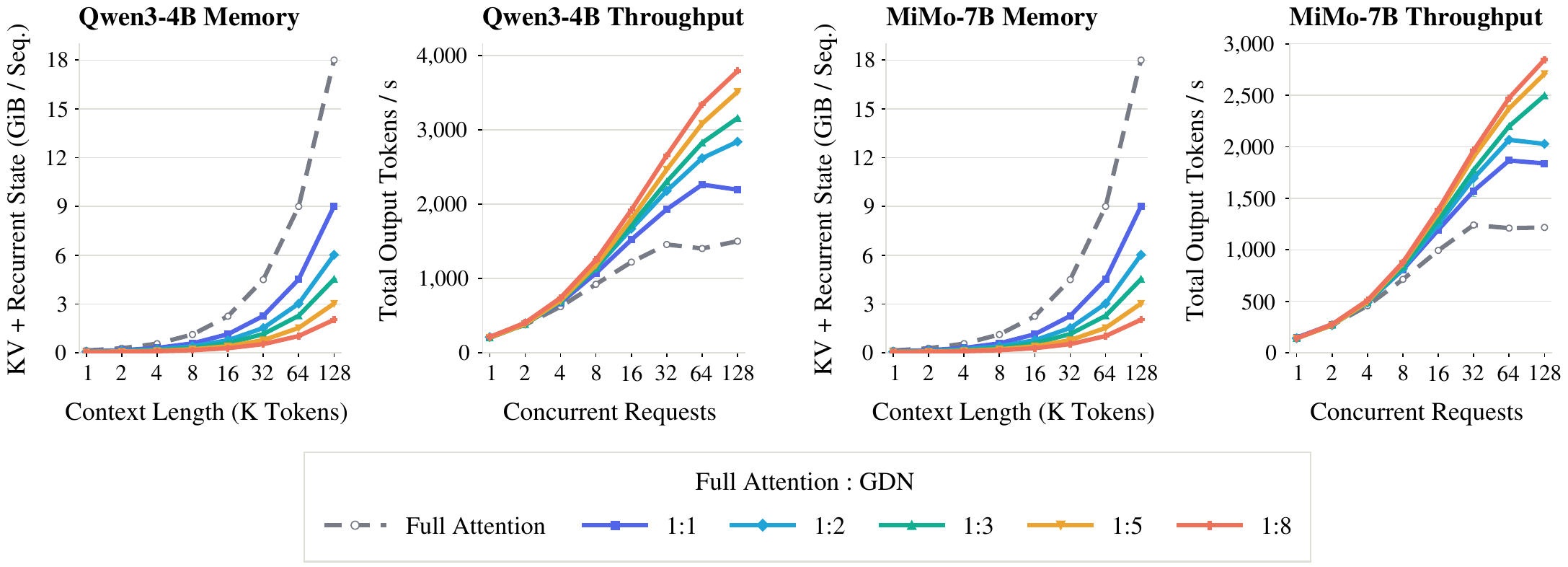}
  \caption{Memory and throughput improvements of GDN hybrids over the full attention teacher.}
  \label{fig:efficiency}
  \vspace{-4mm}
\end{figure}

\textbf{Reasoning Scales with OPD Compute.} Figure~\ref{fig:math_scaling} shows how mathematical reasoning performance evolves with additional OPD training. Performance improves substantially as the number of OPD tokens increases across both models. On MATH-500, most of the improvement occurs early, after which performance saturates near the teacher. In contrast, AIME performance continues to improve throughout training with the strongest checkpoints occurring near the end of our 2B token training budget. This suggests that OPD compute is especially valuable for more difficult reasoning tasks and, since AIME has not clearly saturated, that further scaling may yield additional gains.

\textbf{Comparison to Off-Policy Controls.} To isolate the benefit of training on student-generated trajectories, we compare OPD against matched off-policy baselines. In these controls, the post-Stage 2 hybrid student is trained on the Stage 3 data mixture but using off-policy KD. 
\begin{wraptable}{r}{0.60\textwidth}
    \vspace{-5.5mm}
    \centering
    \caption{OPD versus off-policy controls at matched training budgets. Parenthesis show source of training traces.}
    \label{tab:opd-offpolicy}
    \small
    \setlength{\tabcolsep}{2.5pt}
    \renewcommand{\arraystretch}{1.05}

    \resizebox{\linewidth}{!}{%
    \begin{tabular}{@{}lcccccc@{}}
        \toprule
        & \multicolumn{2}{c}{\textbf{MATH-500}}
        & \multicolumn{2}{c}{\textbf{AIME'24}}
        & \multicolumn{2}{c}{\textbf{AIME'25}} \\
        \cmidrule(lr){2-3}
        \cmidrule(lr){4-5}
        \cmidrule(lr){6-7}
        \textbf{Training}
        & \textbf{200M} & \textbf{400M}
        & \textbf{200M} & \textbf{400M}
        & \textbf{200M} & \textbf{400M} \\
        \midrule
        OPD (Student)
            & $\mathbf{83.6}$ & $\mathbf{87.0}$
            & $\mathbf{23.3}$ & $\mathbf{33.3}$
            & $\mathbf{18.8}$ & $\mathbf{31.7}$ \\
        Off-Policy (Dataset)
            & $77.3$ & $78.1$
            & $12.5$ & $17.1$
            & $14.6$ & $15.8$ \\
        Off-Policy (Teacher)
            & $75.5$ & $80.0$
            & $12.9$ & $17.5$
            & $13.8$ & $16.7$ \\
        \bottomrule
    \end{tabular}%
    }
    \renewcommand{\arraystretch}{1.0}
    \vspace{-2mm}
\end{wraptable}
The traces used for training are either the dataset ground-truth text, as in Stage 2, or teacher traces generated by the full attention model. Each experiment uses Qwen3-4B as the model backbone. The off-policy baselines significantly underperform relative to OPD. At 400M tokens, OPD improves over the extended Stage 2 recipe by $16.2\%$ on AIME'24 and $15.9\%$ on AIME'25. Training on teacher-generated traces marginally improves performance, but this regime still lags behind OPD by up to $15.8\%$. The results support the hypothesis that OPD is necessary for correcting errors induced by the recurrent state in hybrid models over long contexts.

\subsection{Additional Experiments}
\textbf{Efficiency Gains.} Figure~\ref{fig:efficiency} evaluates the inference efficiency gains from attention linearization. We serve both the full attention and hybrid models using vLLM. At the 1:3 full attention to GDN ratio used
by \AlgName, cache and recurrent-state memory is approximately $4\times$ lower than full attention at 128K context, where KV cache memory dominates overall memory. At the highest concurrency, throughput is approximately $2\times$ higher for both Qwen and MiMo hybrids. Thus, \AlgName\ preserves the efficiency benefits that motivate hybrid attention architectures in the first place.
\begin{figure}[t]
  \centering
  \includegraphics[width=\linewidth]{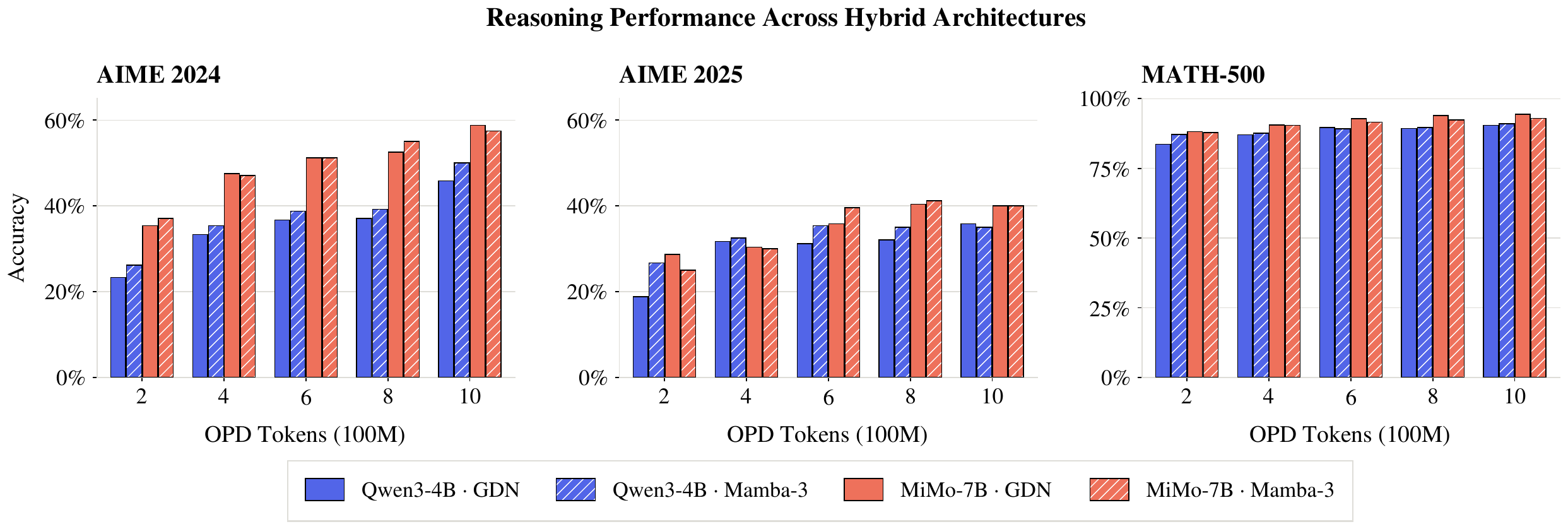}
  \caption{Comparison of how reasoning scales for various hybrid architectures.}
  \vspace{-6mm}
  \label{fig:mamba}
\end{figure}

\textbf{Alternative Linear Attention Architectures.} We apply \AlgName\ to Qwen3-4B and MiMo-7B-RL-0530 using Mamba-3 \citep{lahoti2026mamba}. Figure~\ref{fig:mamba} shows that Mamba-3 hybrids achieve reasoning performance competitive with GDN at matched OPD budgets. At 1B OPD tokens, Qwen3 with Mamba-3 reaches 91.0\%, 50.0\%, and 35.0\% on MATH-500, AIME'24, and AIME'25, respectively. At 800M tokens, MiMo-7B-RL-0530 with Mamba-3 reaches 92.4\%, 55.0\%, and 41.2\%. Thus, the \AlgName\ procedure is not GDN-specific and can applied to models using various efficient attention architectures while maintaining reasoning and long-context capabilities. 

\textbf{Comparison to Non-Distillation Baselines.} We compare \AlgName\ against two non-distillation attention conversions (Table \ref{tab:nondistill}). First, we convert both Qwen3-4B and MiMo-7B-RL-0530 to hybrid sliding window attention (SWA) using a 256-token window and four sinks \citep{Beltagy2020Longformer, jolicoeurmartineau2026slidingwindowbeatslinearattention}. We retain the same 1:3 full attention to converted layer ratio as our GDN hybrids to approximately match long-context memory budgets. For MiMo-7B-RL-0530, we convert the model to multi-head latent attention (MLA) \citep{deepseekai2024deepseekv2strongeconomicalefficient} by following the TransMLA \citep{meng2025transmla} conversion recipe with 3B recovery SFT tokens. The converted SWA and TransMLA models incur substantial losses in reasoning and long-context retrieval. Our OPAL-Qwen3 and OPAL-MiMo hybrids reach $92.0\%$ and $93.8\%$ on MATH-500 versus $17.4\%$ and $51.0\%$ for SWA and $27.0\%$ for MiMo-7B-RL-0530 with TransMLA. On AIME and 32K token context NIAH, our \AlgName\ hybrids outperform both conversion baselines which completely collapse.
\begin{table*}[t]
    \centering

    \caption{Comparison to non-distillation baselines on NIAH (32K) and math reasoning (\%).}
    \label{tab:nondistill}

    \setlength{\tabcolsep}{6pt}
    \renewcommand{\arraystretch}{1.0}

    \resizebox{\textwidth}{!}{\scriptsize%
    \begin{tabular}{lcccccc}
        \toprule
        & \multicolumn{3}{c}{\textbf{NIAH, 32K Context}} & \multicolumn{3}{c}{\textbf{Math Reasoning}} \\
        \cmidrule(lr){2-4}
        \cmidrule(lr){5-7}
        \textbf{Model / Conversion} & \textbf{Single Needle} & \textbf{Multi-Key} & \textbf{Multi-Query} & \textbf{MATH-500} & \textbf{AIME'24} & \textbf{AIME'25} \\
        \midrule
        Qwen3-4B & $100.0$ & $99.8$ & $100.0$ & $96.4$ & $73.8$ & $63.7$ \\
        \quad $\hookrightarrow$ SWA$(256,4)$ & $0.0$ & $1.0$ & $0.2$ & $17.4$ & $0.0$ & $0.0$ \\
        \quad $\hookrightarrow$ GDN (ours, 2B OPD) & $\mathbf{100.0}$ & $\mathbf{99.6}$ & $\mathbf{100.0}$ & $\mathbf{92.0}$ & $\mathbf{52.5}$ & $\mathbf{35.0}$ \\
        \addlinespace
        MiMo-7B-RL-0530 & $100.0$ & $92.8$ & $76.2$ & $97.0$ & $73.3$ & $67.9$ \\
        \quad $\hookrightarrow$ SWA$(256,4)$ & $0.0$ & $1.0$ & $0.3$ & $51.0$ & $1.7$ & $1.7$ \\
        \quad $\hookrightarrow$ TransMLA (3B Recovery) & $39.0$ & $2.0$ & $1.5$ & $27.0$ & $0.4$ & $0.0$ \\
        \quad $\hookrightarrow$ GDN (ours, 2B OPD) & $\mathbf{100.0}$ & $\mathbf{83.8}$ & $\mathbf{100.0}$ & $\mathbf{93.8}$ & $\mathbf{59.6}$ & $\mathbf{47.1}$ \\
        \bottomrule
    \end{tabular}%
    }

    \vspace{-6mm}
\end{table*}

%% file: Sections/Conclusion.tex
\section{Conclusion}
We present \AlgName, a method for converting fully trained, full-attention transformers to hybrid models while maintaining their long-context retrieval and reasoning abilities. Traditional off-policy distillation baselines remain performant on likelihood tasks, but suffer performance degradation on more involved, long-horizon benchmarks. \AlgName\ addresses this failure mode with a prolonged, continual-learning-style OPD stage, providing a training signal for the hybrid student to learn from the states it encounters during generation. This is especially relevant for hybrid students whose compressed recurrent state can accumulate errors over long contexts. When applied to Qwen3-4B and MiMo-7B-RL-0530, \AlgName\ fully recovers long-context retrieval performance and closes the gap on mathematical reasoning while retaining only a quarter of the full-attention layers. Broadly, our findings position on-policy training as an important ingredient for post-hoc architectural conversion, and we aim to apply a similar training recipe to other architectural conversions to show that capabilities do not have to be tied to the architecture in which they were originally learned. 

%% file: Sections/Post_Paper.tex
\section*{Acknowledgments}

This work was partially supported by NSF grants CCF 2045694, CCF 2428569, CCF 2339084, CNS-2112471, CPS-2111751, ONR grant N00014-23-1-2149, and an AI2C Seed grant. This work used Bridges-2 GPU at the Pittsburgh Supercomputing Center through allocations CIS250149 and CIS250011 from the Advanced Cyberinfrastructure Coordination Ecosystem: Services \& Support (ACCESS) program, which is supported by NSF grants \#2138259, \#2138286, \#2138307, \#2137603, and \#2138296.  \citep{access}. We gratefully acknowledge use of the research computing resources of the Empire AI Consortium, Inc, with support from Empire State Development of the State of New York, the Simons Foundation, and the Secunda Family Foundation \citep{Bloom2025EmpireAI}. We would like to thank Raghav Singhal and Kaustubh Ponkshe for providing valuable comments and feedback on the manuscript. 

%% file: Sections/Appendix.tex
\appendix
\section*{Appendix}
\input{Sections/Appendix_Sections/Related_Works}
\input{Sections/Appendix_Sections/Experimental_Details}

\input{Sections/Appendix_Sections/Additional_Experiments}

%% file: Sections/Appendix_Sections/Related_Works.tex
\section{Related works}
\textbf{Linear Attention and Hybrid Models.} Linear attention and recurrent sequence models replace softmax attention with a fixed-size state rather than retaining the full history of key-value pairs, enabling subquadratic sequence processing \citep{pmlr-v119-katharopoulos20a,gu2024mamba,dao2024mamba2,yang2025gated}. However, pure fixed-state models exhibit limitations in recall and retrieval \citep{arora2024zoology, jelassi2024repeat}. While these architectures are competitive at smaller scales, many large-scale efficient language models still retain some form of full attention, motivating hybrid architectures \citep{lenz2025jamba,kimiteam2026kimik3openfrontier,qwen38}. Hybrid models introduce interesting design choices, including the ratio and placement of full and linear attention layers \citep{lenz2025jamba}, the choice of recurrent or linear attention mixer, and whether the two mechanisms are interleaved across layers or combined within the same layer \citep{dong2025hymba}. A complementary design is explored by Samba \citep{ren2025samba}, which combines Mamba with sliding window attention rather than relying on a small number of global full-attention layers.

\textbf{Architecture Conversion after Pretraining.} Architectural modification after pretraining is an emerging approach to improving model efficiency without paying the cost of retraining from scratch. At the lightest end of this spectrum, new modules can be added while leaving the pretrained backbone largely or entirely unchanged, such as auxiliary decoding heads for speculative generation \citep{cai2024medusa,li2024eagle} or gated cross attention layers that adapt a frozen language model to multimodal inputs \citep{alayrac2022flamingo}. More substantial conversions modify the pretrained backbone itself. Dense models can be upcycled into sparse mixture-of-experts (MoE) models by converting dense FFNs into experts and continuing pretraining \citep{komatsuzaki2023sparseupcycling}, while autoregressive language models can be adapted into diffusion models by relaxing causal attention to bidirectional attention \citep{gong2025diffullama}. The attention mechanism itself can be replaced after pretraining, including MHA-to-GQA and GQA-to-MLA conversion at a small fraction of the original pretraining cost \citep{ainslie2023gqa,meng2025transmla}. Pretrained layers can similarly be tied and reused recurrently to trade parameters for additional computational depth \citep{bae2025relaxed,shapiro2026retrofitting}, while structured pruning followed by continued training or off-policy distillation can produce smaller models while recovering much of the original performance \citep{xia2024sheared,muralidharan2024minitron}. These approaches are evaluated on different benchmarks and sometimes do not recover long-context performance despite involving less disruptive architectural changes. Methods targeting capability recovery have generally relied on significant continued pretraining.

\textbf{Hybridizing Full-Attention Transformers.} Existing methods for converting pretrained full-attention transformer layers into linear attention layers typically use some combination of layerwise alignment (Stage 1), end-to-end off-policy distillation (Stage 2), and continual pretraining on additional data. Lightweight approaches primarily align each converted layer with the attention layer it replaces, learning feature maps or recurrent parameters while largely preserving the pretrained weights \citep{kasai2021t2r,zhang2024hedgehog,zhang2025lolcats,lan2025liger}. A second family supplements such initialization with end-to-end off-policy distillation, matching the converted student's predictions to a frozen full-attention teacher on fixed sequences \citep{bick2024mohawk,wang2024mambainllama,bick2025llamba,goldstein2025radlads,yang2025zebrallama,chen2026hybridlinearattentionright,li2026distilling,bick2026retrievalaware,lan2026morphinghybridattentionmodels}. Other approaches instead recover the converted model through substantial additional next-token training \citep{mercat2024supra,chattopadhyay2026priming}. While all these methods typically recover short-context quality effectively, long-context retrieval often lags behind the source transformer while long-context reasoning/generation typically collapses. Priming \citep{chattopadhyay2026priming} is a notable exception, demonstrating strong long-context reasoning with a 1:1 full attention:GDN/SSM-attention hybrid, but requires multi-stage continual pretraining at context lengths up to 128K, supervised fine-tuning on chain-of-thought traces, and instruction alignment while using 150B tokens, approaching the Chinchilla compute-optimal \citep{hoffmann2022training} pretraining budget (160B) for the 8B parameter model considered.

\textbf{On-Policy Distillation and Long-Horizon Reasoning.} Reasoning models often generate long trajectories, making the distribution of model-generated prefixes increasingly important as generation proceeds. Recent progress in reasoning has been driven largely by reinforcement learning with verifiable rewards (RLVR), with GRPO and its variants providing a scalable means of optimizing directly on model-generated trajectories \citep{shao2024deepseekmath,guo2025deepseekr1,yu2025dapo}. On-policy distillation has emerged as a complementary approach, replacing sparse outcome rewards with dense token-level supervision from a stronger teacher on the student's own generations \citep{agarwal2024onpolicy,lu2025onpolicydistillation}. This has been particularly effective for transferring reasoning capabilities from larger to smaller models \citep{qwen2025qwen3}. More recently, multi-teacher on-policy distillation has been used to integrate independently trained reasoning capabilities in frontier-scale post-training \citep{ma2026mopd,mimo2026v2flash}. Notably, Nemotron 3 Ultra, a 550B-parameter MoE hybrid Mamba hybrid model, combines SFT, RL, and multi-teacher on-policy distillation in its post-training pipeline \citep{nvidia2026nemotron3ultra}, demonstrating that on-policy distillation is also applicable to large hybrid reasoning models.

%% file: Sections/Appendix_Sections/Experimental_Details.tex
\newcommand{\sci}[2]{$#1\!\times\!10^{#2}$}

\section{Experimental Details}
\subsection{Models and Hybrid Construction}
\begin{table}[!ht]
\vspace{-4mm}
\centering
\footnotesize
\setlength{\tabcolsep}{4pt}
\caption{Teacher models and the converted hybrids. Both retain the same
nine full attention layers and use Gated DeltaNet (GDN) as the linear attention layer.}
\label{tab:arch}
\begin{tabular}{@{}lll@{}}
\toprule
 & Qwen3-4B  & MiMo-7B-RL-0530  \\
\midrule
Teacher          & \texttt{Qwen/Qwen3-4B} & \texttt{XiaomiMiMo/MiMo-7B-RL-0530} \\
Teacher Parameters        & 4.02B & 7.62B (MTP head dropped) \\
Layers        & 36, GQA 32 Q/8 KV, $d_h=128$ & 36, GQA 32 Q/8 KV, $d_h=128$ \\
Context Length            & 32,768  & 65,536 \\
Retained Full Attention Layers  & \multicolumn{2}{l}{9 (Layers $3,7,\dots,35$)} \\
Converted Layers          & \multicolumn{2}{l}{27 (Layers 0, 1, 2, 4, 5, 6, \dots, 34)} \\
GDN Initialization        & \multicolumn{2}{l}{Q/K/V/O inherited from full attention teachers. Others set to default values} \\
Hybrid Student Parameters         & 4.55B (1.14B in GDN Layers) & 8.31B (1.82B in GDN Layers) \\
\bottomrule
\end{tabular}
\end{table}

\subsection{Training Procedure and Hyperparameters}
All runs were conducted using 3–4 NVIDIA H100 or H200 GPUs, depending on availability.

\begin{table*}[!ht]
\vspace{-4mm}
\centering
\small
\setlength{\tabcolsep}{5pt}
\caption{Optimization hyperparameters by stage. All stages use AdamW
($\beta_1{=}0.9$, $\beta_2{=}0.95$, weight decay 0), gradient clipping at
norm 1.0. ``Inherited'' denotes all
non-GDN weights.}
\label{tab:hparams}
\begin{tabular}{@{}l cc cc cc cc@{}}
\toprule
 & \multicolumn{2}{c}{Stage 1} & \multicolumn{2}{c}{Stage 2A} & \multicolumn{2}{c}{Stage 2B} & \multicolumn{2}{c}{Stage 3} \\
\cmidrule(lr){2-3}\cmidrule(lr){4-5}\cmidrule(lr){6-7}\cmidrule(l){8-9}
 & Qwen & MiMo & Qwen & MiMo & Qwen & MiMo & Qwen & MiMo \\
\midrule
Trainable     & GDN & GDN & All & All & All & All & All & All \\
Tokens         & 100M & 100M & 600M & 600M & 294M & 294M & 2B & 2B \\
Sequence length          & 4K & 4K & 4K & 4K & 32K & 32K & 32K & 32K \\
Tokens per batch          & 32K & 16K & 2.1M & 2.1M & 2.1M & 2.1M & 131K & 131K \\
Optimizer Steps          & 3.1K & 6.1K & 287 & 287 & 140 & 140 & 12K & 12K \\
Peak LR, GDN             & \sci{3}{-3} & \sci{1}{-3} & \sci{2}{-4} & \sci{5}{-5} & \sci{1}{-4} & \sci{2.5}{-5} & \sci{2}{-5} & \sci{1}{-5} \\
Peak LR, Inherited       & -- & -- & \sci{2}{-5} & \sci{5}{-6} & \sci{1}{-5} & \sci{2.5}{-6} & \sci{2}{-5} & \sci{1}{-5} \\
Warmup Tokens                & 3M & 3M & 18M & 18M & 8.8M & 8.8M & 20M & 20M \\
LR Schedule                    & Cosine & Cosine & Cosine & Cosine & Cosine & Cosine & WSD & WSD \\
\bottomrule
\end{tabular}
\end{table*}

\begin{table}[!ht]
\vspace{-4mm}
\centering
\small
\setlength{\tabcolsep}{4pt}
\caption{Stage-3 on-policy distillation settings. Rollouts are sampled with
vLLM at $T{=}1.0$, top-$p{=}1.0$ under the thinking-mode chat template.}
\label{tab:opd}
\begin{tabular}{@{}lcc@{}}
\toprule
 & Qwen3-4B & MiMo-7B-RL-0530 \\
\midrule
Generated Tokens per Step        & 131,072 & 131,072 \\
Prompt + Rollout Limit           & 40,960 & 65,536 \\
Initial Horizon $H$              & 512 & 512 \\
Horizon Doubling Rule            & \multicolumn{2}{c}{KL Plateau: rel.\ gain $<$2\% for 30 steps} \\
Maximum Horizon                  & 32K & 32K \\
Sampler Batch (Prompts)          & 64; 32 at $H{=}32$k & 64; 32 at $H{=}32$k \\
Peak LR             & \sci{2}{-5} & \sci{1}{-5} \\
Warmup                           & 150 Steps (20M Tokens) & 150 Steps (20M Tokens) \\
\bottomrule
\end{tabular}
\end{table}

\subsection{OPD Scaling Ladder}
\begin{table}[!ht]
\vspace{-4mm}
\centering
\small
\setlength{\tabcolsep}{6pt}
\caption{OPD scaling ladder configuration. Each rung continues the previous one from its pre-decay
snapshot at the constant learning rate, then decays linearly to the final LR
over the decay window. }
\label{tab:opd-ladder}
\begin{tabular}{@{}lccc@{}}
\toprule
Rung & Max Output Horizon & Decay Start & Decay Tokens \\
\midrule
\multicolumn{4}{@{}l}{\textit{Qwen3-4B}} \\
0 $\rightarrow$ 200M      & 512 $\rightarrow$ 16K & 161M & 39M \\
200M $\rightarrow$ 400M   & 16K & 340M & 60M \\
400M $\rightarrow$ 500M   & 16K & 430M & 70M \\
500M $\rightarrow$ 600M   & 16K & 520M & 80M \\
600M $\rightarrow$ 800M   & 16K $\rightarrow$ 32K & 710M & 90M \\
800M $\rightarrow$ 900M   & 32K & 800M & 100M \\
900M $\rightarrow$ 1.0B   & 32K & 890M & 110M \\
1.0B $\rightarrow$ 1.2B   & 32K & 1.09B & 110M \\
1.2B $\rightarrow$ 1.4B   & 32K & 1.29B & 110M \\
1.4B $\rightarrow$ 1.6B   & 32K & 1.49B & 110M \\
1.6B $\rightarrow$ 1.8B   & 32K & 1.69B & 110M \\
1.8B $\rightarrow$ 2.0B   & 32K & 1.89B & 110M \\
\addlinespace \addlinespace
\multicolumn{4}{@{}l}{\textit{MiMo-7B-RL-0530}} \\
0 $\rightarrow$ 200M      & 512 $\rightarrow$ 16K & 161M & 39M \\
200M $\rightarrow$ 400M & 16K & 340M & 63M \\
400M $\rightarrow$ 600M   & 16K & 537M & 63M \\
600M $\rightarrow$ 800M   & 16K $\rightarrow$ 32K & 737M & 63M \\
800M $\rightarrow$ 1.0B   & 32K & 937M & 63M \\
1.0B $\rightarrow$ 1.2B   & 32K & 1.14B & 63M \\
1.2B $\rightarrow$ 1.4B   & 32K & 1.34B & 63M \\
1.4B $\rightarrow$ 1.6B   & 32K & 1.52B & 79M \\
1.6B $\rightarrow$ 1.8B   & 32k & 1.72B & 79M \\
1.8B $\rightarrow$ 2.0B   & 32k & 1.92B & 79M \\
\bottomrule
\end{tabular}
\end{table}

%% file: Sections/Appendix_Sections/Additional_Experiments.tex
\section{Additional Experiments}
\subsection{Commonsense Reasoning Performance}

\begin{table}[!ht]
    \vspace{-4mm}
    \centering
    \caption{Commonsense reasoning performance (\%). \TeacherTag; \StudentTag.}
    \label{tab:commonsense}
    \setlength{\tabcolsep}{6pt}
    \renewcommand{\arraystretch}{1.0}
    \resizebox{\textwidth}{!}{%
        \begin{tabular}{lccccccc}
            \toprule
            \textbf{Model}
            & \textbf{PIQA}
            & \textbf{HellaSwag}
            & \textbf{ARC-E}
            & \textbf{ARC-C}
            & \textbf{Winogrande}
            & \textbf{MMLU}
            & \textbf{Avg.} \\
            \midrule

            \teacherrow
            \textit{Qwen2.5-7B-Instruct}
                & $80.2$ & $80.5$ & $80.9$ & $55.0$ & $70.5$ & $74.3$
                & $73.6$ \\
            \studentrow
            QRWKV6-7B-Instruct
                & $\mathbf{79.9}$ & $78.9$ & $\mathbf{79.3}$ & $55.9$ & $71.4$ & $\mathbf{64.2}$
                & $\mathbf{71.6}$ {\scriptsize ($97.3\%$)} \\

            \midrule
            \teacherrow
            \textit{Llama-3.2-1B-Instruct}
                & $74.9$ & $61.6$ & $63.9$ & $37.6$ & $61.4$ & $46.0$
                & $57.6$ \\
            \studentrow
            Llamba-1B
                & $73.7$ & $61.8$ & $65.2$ & $37.5$ & $61.9$ & $31.5$
                & $55.3$ {\scriptsize ($96.0\%$)} \\

            \midrule
            \teacherrow
            \textit{Llama-3.2-3B-Instruct}
                & $76.8$ & $71.6$ & $71.0$ & $46.4$ & $68.7$ & $60.7$
                & $65.9$ \\
            \studentrow
            Llamba-3B
                & $78.0$ & $74.0$ & $73.6$ & $46.2$ & $\mathbf{71.7}$ & $50.1$
                & $65.6$ {\scriptsize ($99.6\%$)} \\
            \studentrow
            Zebra-Llama-3B
                & $77.3$ & $72.7$ & $75.8$ & $52.3$ & $65.7$ & $51.7$
                & $65.9$ {\scriptsize ($100.1\%$)} \\

            \midrule
            \teacherrow
            \textit{Llama-3.1-8B-Instruct}
                & $81.4$ & $79.5$ & $79.9$ & $55.6$ & $73.6$ & $68.4$
                & $73.1$ \\
            \studentrow
            Zebra-Llama-8B
                & $\mathbf{79.9}$ & $\mathbf{79.0}$ & $77.5$ & $\mathbf{57.6}$ & $71.3$ & $59.1$
                & $70.7$ {\scriptsize ($96.8\%$)} \\

            \midrule
            \teacherrow
            \textit{Qwen3-1.7B}
                & $72.0$ & $60.4$ & $69.5$ & $43.0$ & $61.6$ & $60.2$
                & $61.1$ \\
            \studentrow
            HypeNet-2B
                & $72.0$ & $57.2$ & $67.3$ & $42.7$ & $61.7$ & $39.5$
                & $56.7$ {\scriptsize ($92.8\%$)} \\

            \midrule
            \teacherrow
            \textit{Qwen3-4B}
                & $74.9$ & $68.5$ & $78.5$ & $53.8$ & $65.8$ & $70.1$
                & $68.6$ \\
            \studentrow
            HypeNet-5B
                & $76.3$ & $66.1$ & $75.9$ & $50.6$ & $67.2$ & $52.4$
                & $64.8$ {\scriptsize ($94.4\%$)} \\
            \studentrow
            OPAL-Qwen3 (ours)
                & $72.3$ & $58.5$ & $65.0$ & $41.6$ & $63.0$ & $57.2$
                & $59.6$ {\scriptsize ($86.9\%$)} \\

            \midrule
            \teacherrow
            \textit{MiMo-7B-RL-0530}
                & $74.5$ & $64.9$ & $68.1$ & $45.0$ & $61.6$ & $57.2$
                & $61.9$ \\
            \studentrow
            OPAL-MiMo (ours)
                & $72.5$ & $59.9$ & $62.7$ & $39.5$ & $58.8$ & $54.6$
                & $58.0$ {\scriptsize ($93.7\%$)} \\

            \bottomrule
        \end{tabular}%
    }
\end{table}
\clearpage
\subsection{WSD Training Curves}

\begin{figure}[!hb]
  \centering
  \includegraphics[width=0.76\linewidth]{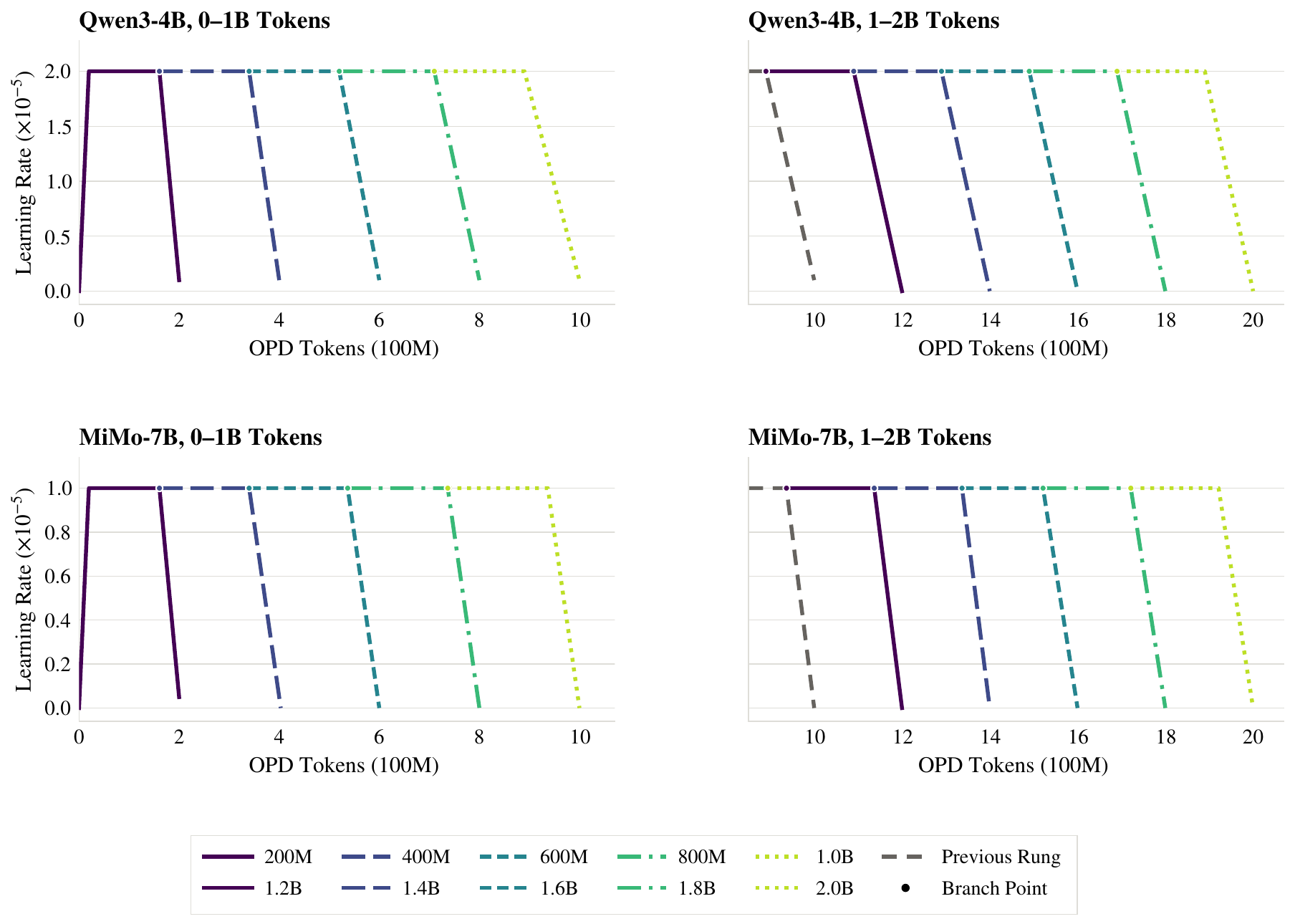}
  \vspace{-3mm}
  \caption{Training horizons for each rung of the OPD scaling ladder. Decay horizons gradually become longer at subsequent rungs before plateauing in the last few rungs.}
  \label{fig:opd_scaling}
\end{figure}

\begin{figure}[!hb]
  \centering
  \includegraphics[width=0.76\linewidth]{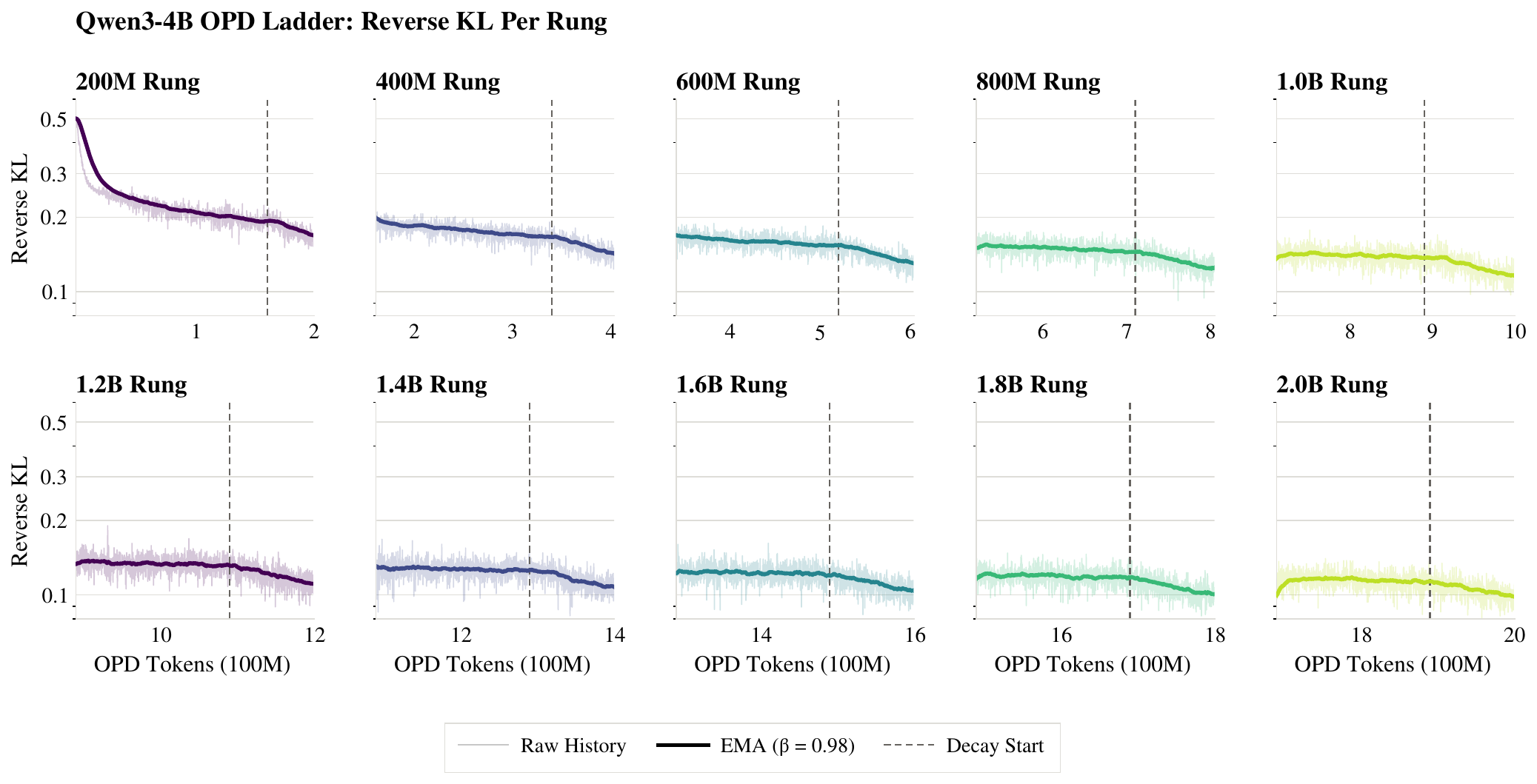}
  \vspace{-3mm}
  \caption{Training curves for Qwen3-4B at each rung using WSD schedule.}
  \label{fig:qwen_wsd}
\end{figure}

\begin{figure}[!hb]
  \centering
  \includegraphics[width=0.76\linewidth]{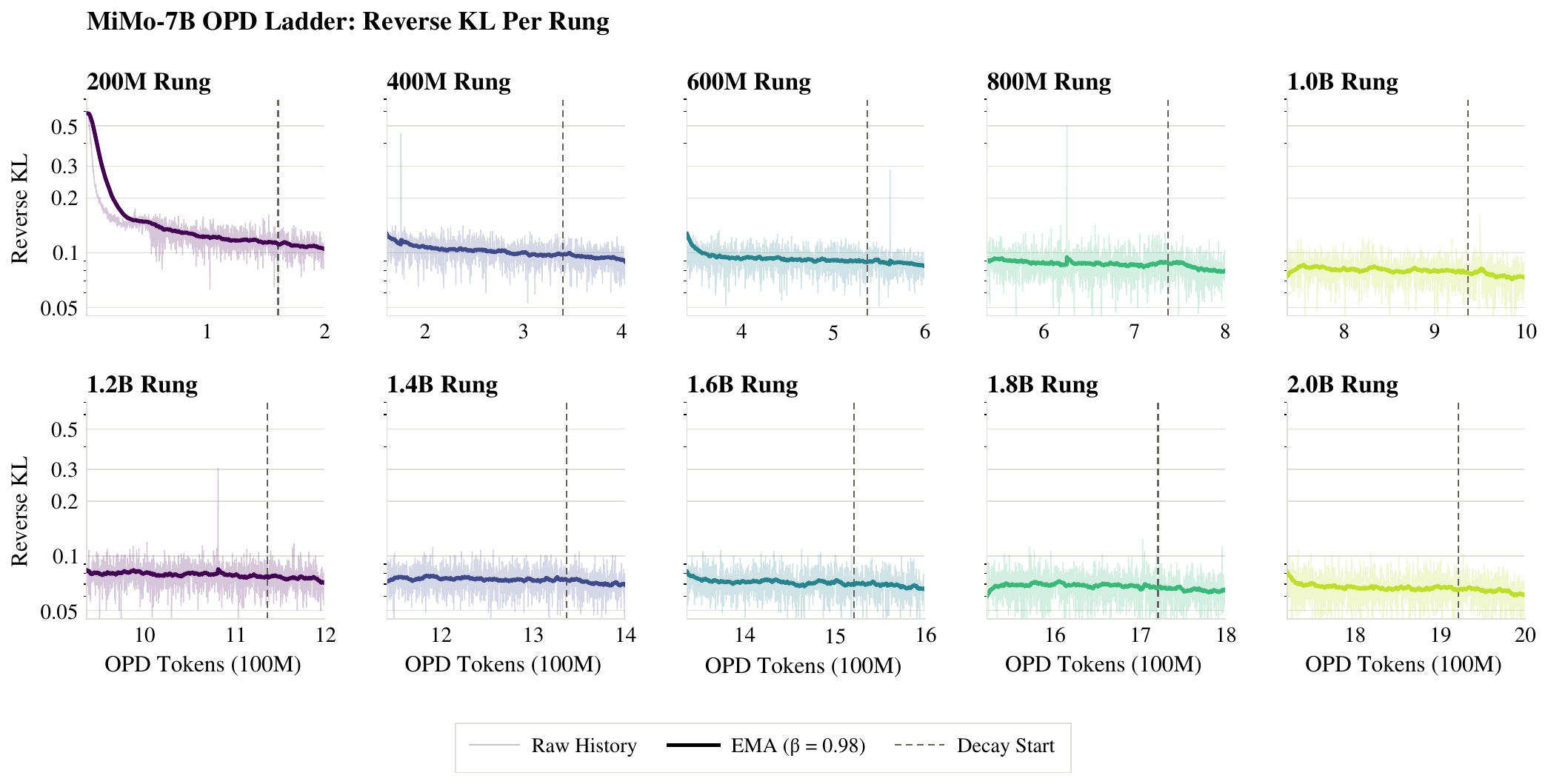}
  \vspace{-3mm}
  \caption{Training curves for MiMo-7B-RL-0530 at each rung using WSD schedule.}
  \label{fig:mimo_wsd}
\end{figure}

\clearpage

\begin{figure}[!t]
  \centering
  \includegraphics[width=0.80\linewidth]{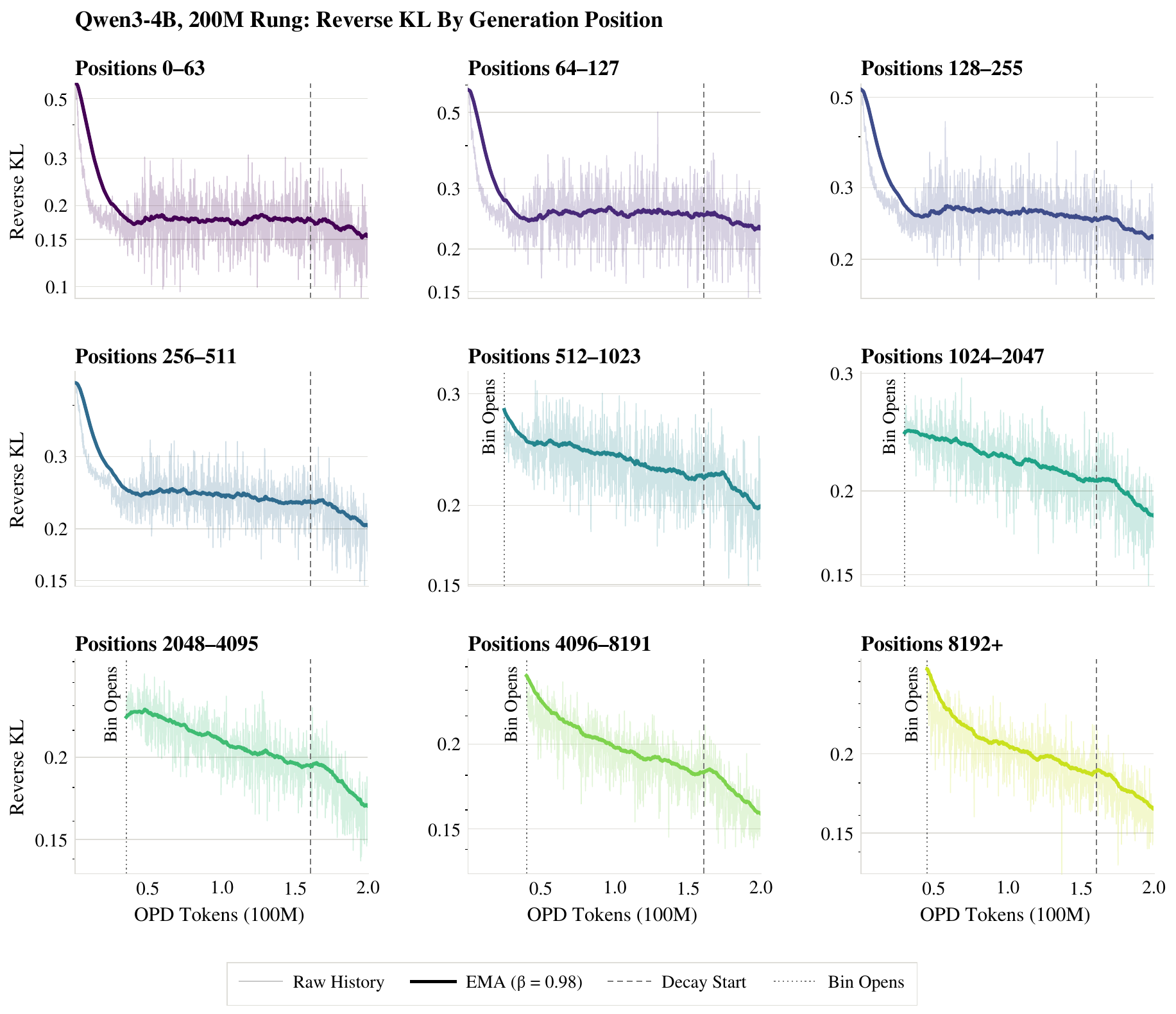}
  \vspace{-3mm}
  \caption{Training curves for position-wise reverse-KL at the first rung of Qwen3-4B training.}
  \label{fig:qwen_wsd_200}
\end{figure}

\begin{figure}[!t]
  \centering
  \includegraphics[width=0.80\linewidth]{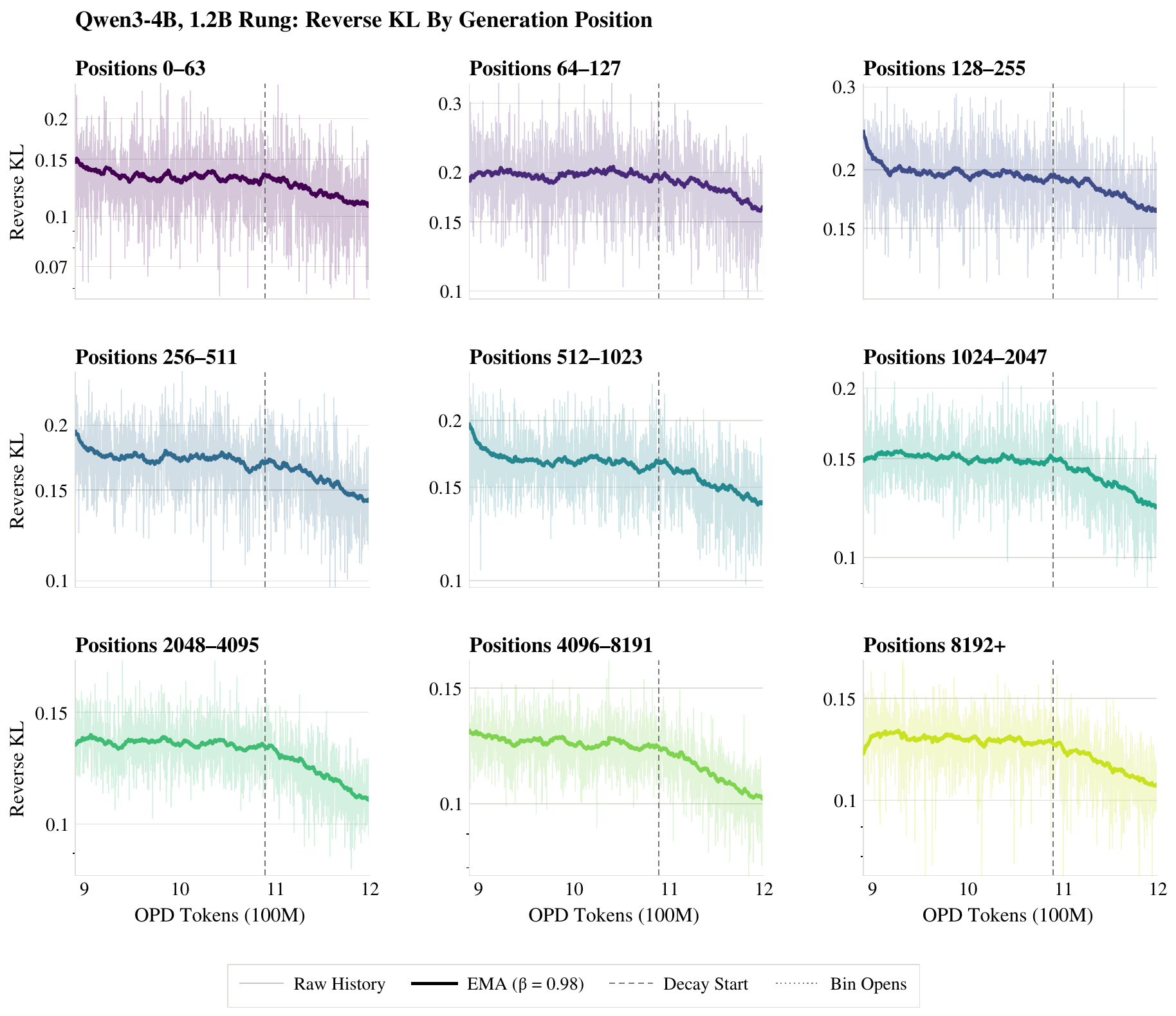}
  \vspace{-3mm}
  \caption{Training curves for position-wise reverse-KL at the run 1B-1.2B OPD tokens of Qwen3-4B training.}
  \label{fig:qwen_wsd_1p2}
\end{figure}

\clearpage

\begin{figure}[!t]
  \centering
  \includegraphics[width=0.80\linewidth]{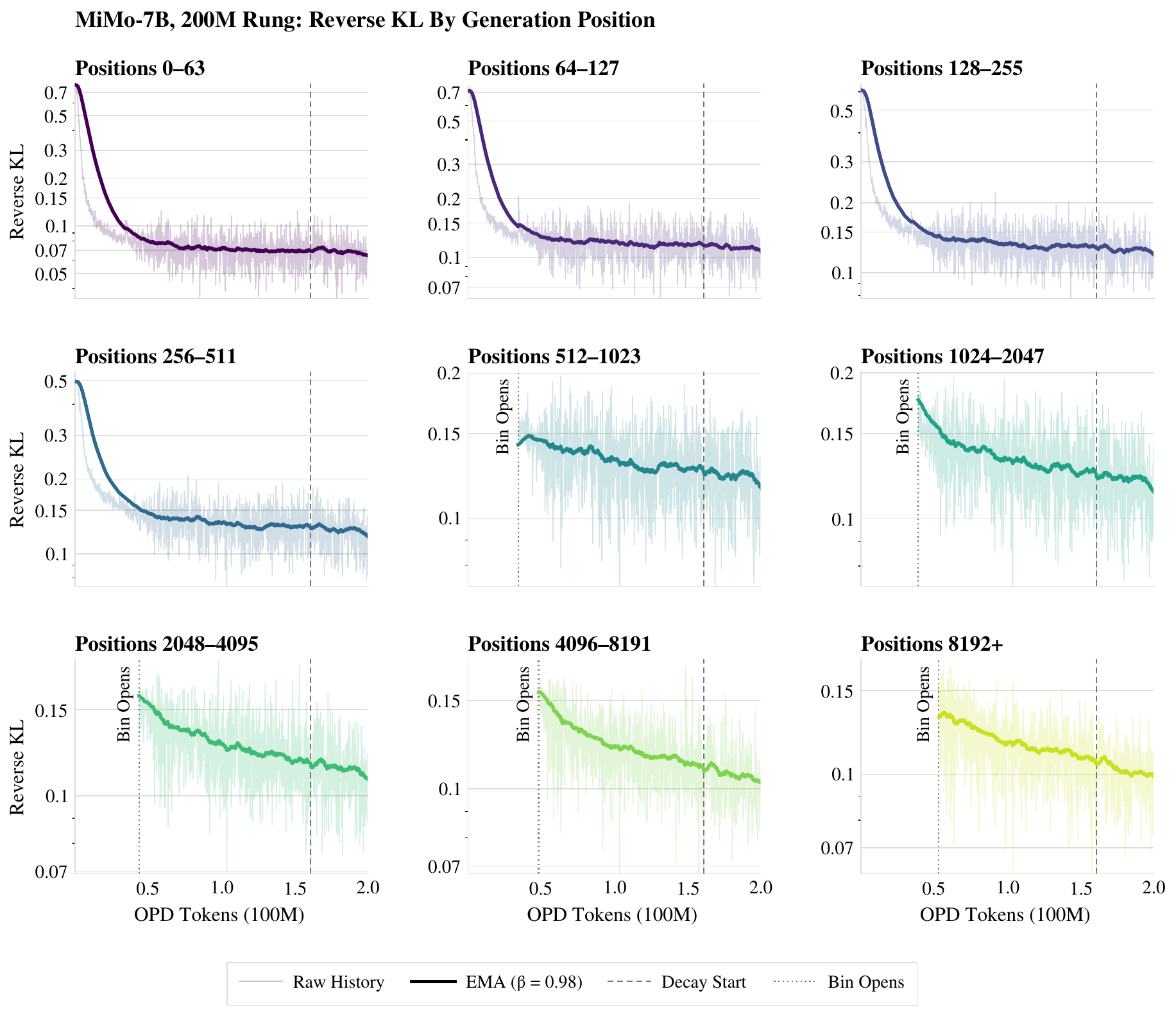}
  \vspace{-3mm}
  \caption{Training curves for position-wise reverse-KL at the first rung of MiMo-7B-RL-0530 training.}
  \label{fig:mimo_wsd_200}
\end{figure}

\begin{figure}[!t]
  \centering
  \includegraphics[width=0.80\linewidth]{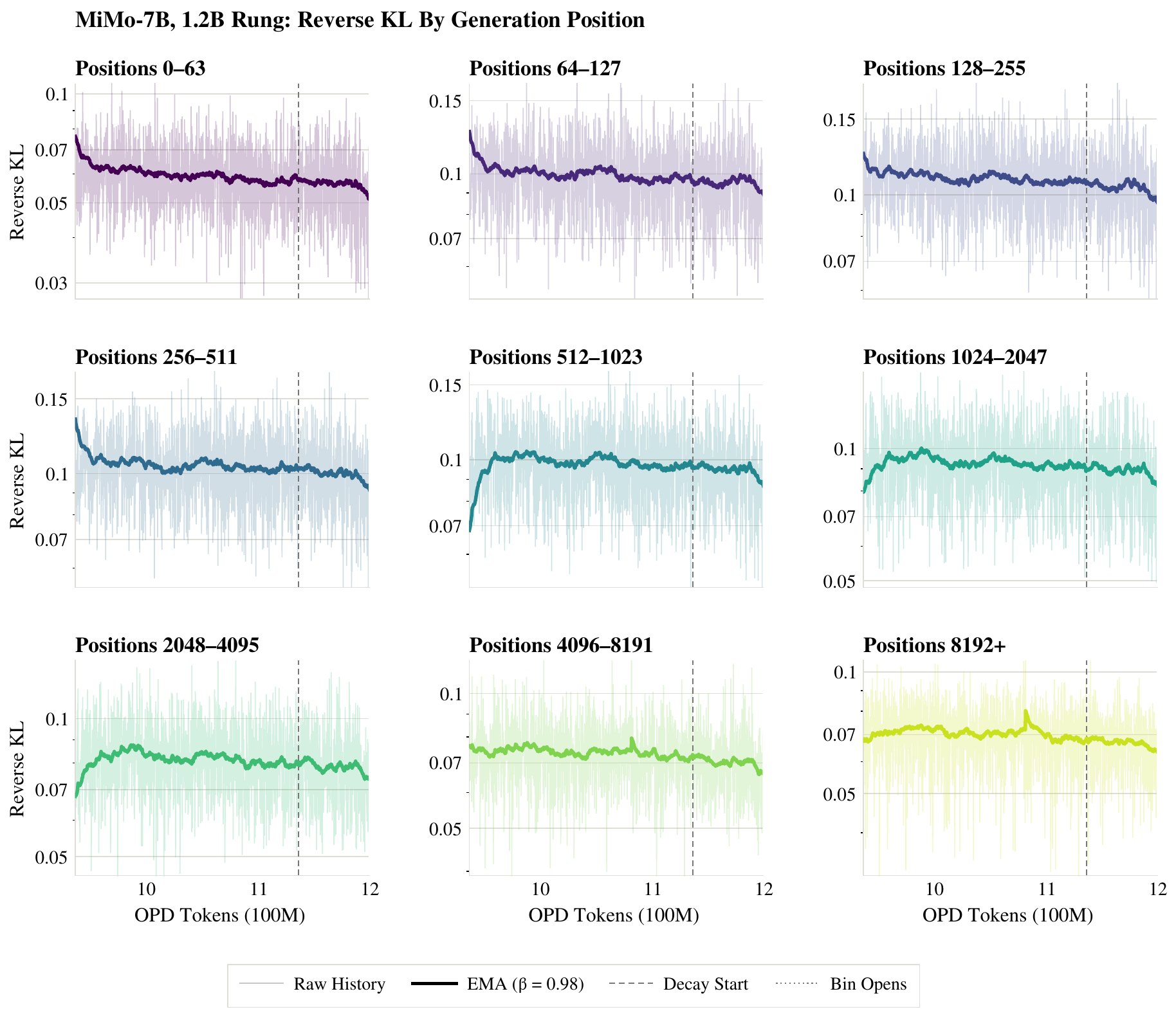}
  \vspace{-3mm}
  \caption{Training curves for position-wise reverse-KL at the run 1B--1.2B OPD tokens of MiMo-7B-RL-0530 training.}
  \label{fig:mimo_wsd_1p2}
\end{figure}

\clearpage
\subsection{Sample Responses}

\newtcolorbox{aimebox}[1]{enhanced, breakable,
  colback=green!4, colframe=green!45!black,
  boxrule=0.5pt, arc=2pt, left=6pt, right=6pt, top=4pt, bottom=4pt,
  title={#1}, fonttitle=\bfseries\small, coltitle=white,
  colbacktitle=green!55!black,
  before skip=8pt, after skip=8pt}

\newtcolorbox{niahbox}[1]{enhanced, breakable,
  colback=blue!4, colframe=blue!45!black,
  boxrule=0.5pt, arc=2pt, left=6pt, right=6pt, top=4pt, bottom=4pt,
  title={#1}, fonttitle=\bfseries\small, coltitle=white,
  colbacktitle=blue!55!black,
  before skip=8pt, after skip=8pt}
\newcommand{\samplelabel}[1]{\par\smallskip\noindent\textbf{\small #1}\par\smallskip}
\input{Sections/Appendix_Sections/Samples/qwen_aime}
\input{Sections/Appendix_Sections/Samples/qwen_niah}
\input{Sections/Appendix_Sections/Samples/mimo_aime}
\input{Sections/Appendix_Sections/Samples/mimo_niah}

%% file: Sections/Appendix_Sections/Samples/qwen_aime.tex
\begin{aimebox}{AIME 2024 I, Problem 13 --- Qwen3-4B GDN Hybrid}
\samplelabel{Problem}
Let $p$ be the least prime number for which there exists a positive integer $n$ such that $n^{4}+1$ is divisible by $p^{2}$. Find the least positive integer $m$ such that $m^{4}+1$ is divisible by $p^{2}$.\hfill\textit{Answer: 110}
\samplelabel{Model reasoning (thinking mode)}
\begin{Verbatim}[breaklines=true, breakanywhere=true, fontsize=\footnotesize, breaksymbolleft={}]
Okay, so I need to find the least prime number p such that there's a positive integer n where n^4 + 1 is divisible by p^2. Then, once I find that p, I have to find the smallest positive integer m where m^4 + 1 is divisible by p^2. Alright, let's start by understanding the problem.

First, the question is about divisibility by p^2. So, for a prime p, we need to find n such that n^4 == -1 (mod p^2). Then, find the smallest such prime p, and then find the minimal m for that p.

So, step one: find the smallest prime p where there exists an integer n with n^4 == -1 (mod p^2).

To approach this, maybe I can start by checking small primes and see if they satisfy the condition.

First, check p=2. Let's see if there's an n such that n^4 + 1 is divisible by 4.

n^4 mod 4: Let's compute n^4 mod 4 for n=1,2,3,...

n=1: 1^4 = 1 mod 4, so 1 + 1 = 2 mod 4 != 0.

n=2: 16 mod 4 = 0, so 0 + 1 = 1 mod 4 != 0.

n=3: 81 mod 4 = 1, so 1 + 1 = 2 mod 4 != 0.

n=4: 256 mod 4 = 0, same as n=2. So n^4 + 1 is 1 mod 4. So no solution for p=2.

Next prime is p=3.

Check if there exists n such that n^4 + 1 == 0 (mod 9).

So n^4 == -1 (mod 9).

Let me compute n^4 mod 9 for n=1 to 8.

n=1: 1

n=2: 16 mod9=7

n=3: 81 mod9=0

n=4: 256 mod9. Let's compute 256 /9: 9*28=252, 256-252=4, so 4.

n=5: 625 mod9. 625 /9=69*9=621, 625-621=4. So 4.

n=6: 1296 mod9=0.

[... 9,743 tokens of reasoning omitted ...]

But since we already confirmed that 110^4 == -1 (mod 289), then 110^4 + 1 is divisible by 289. Therefore, yes. Therefore, 110 is the answer.

Therefore, I think the answer is 110. However, I need to check if there is a mistake in my reasoning. For example, maybe there is a smaller prime p=17, but for which m is smaller than 110. But according to our previous check, no. Therefore, I think the answer is 110.

Wait, but let me check if m= 110 is indeed the minimal. Let me check m= 110 - 289= -179, which is negative, so not allowed. Therefore, yes, 110 is the minimal positive integer.

Therefore, the answer is \boxed{110}.

**Final Answer**
\boxed{110}
\end{Verbatim}
\samplelabel{Model answer}
\begin{Verbatim}[breaklines=true, breakanywhere=true, fontsize=\footnotesize, breaksymbolleft={}]
To solve the problem, we are given the condition that for a prime number $ p $, there exists a positive integer $ n $ such that $ n^4 + 1 $ is divisible by $ p^2 $. Our task is to:

1. **Find the smallest such prime $ p $** for which this is true.
2. **Find the smallest positive integer $ m $** such that $ m^4 + 1 $ is divisible by $ p^2 $, where $ p $ is the smallest prime from step 1.

---

### Step 1: Understanding the Condition

We are looking for the **smallest prime** $ p $ such that there exists an integer $ n $ with:
$$
n^4 + 1 \equiv 0 \pmod{p^2}
$$
This is equivalent to:
$$
n^4 \equiv -1 \pmod{p^2}
$$
This implies that $ -1 $ must be a **quartic residue modulo $ p^2 $**, or more specifically, that there exists an integer $ n $ such that $ n^4 \equiv -1 \mod p^2 $.

---

### Step 2: Modular Conditions on $ p $

Let's analyze the congruence $ n^4 \equiv -1 \mod p $, which is necessary for $ n^4 + 1 \equiv 0 \mod p $.

- $ n^4 \equiv -1 \mod p $ implies that $ -1 $ is a **4th power residue modulo $ p $**.
- This is only possible if $ p \equiv 1 \mod 8 $, because $ -1 $ is a 4th power modulo a prime $ p $ **if and only if** $ p \equiv 1 \mod 8 $.

Thus, the **smallest prime** $ p $ for which this is possible is the **first prime congruent to 1 mod 8**, which is:

$$
p = 17
$$

---

### Step 3: Verifying $ p = 17 $

We now check whether there exists an integer $ n $ such that $ n^4 + 1 $ is divisible by $ 17^2 = 289 $.

We first confirm that $ n^4 \equiv -1 \mod 17 $ has solutions. Since 17 is $ 1 \mod 8 $, it is known that $ -1 $ is a 4th power residue modulo 17. Indeed, we can verify that:

$$
8^4 = 4096 \equiv -1 \mod 17
$$

So $ n = 8 $ is a solution to $ n^4 \equiv -1 \mod 17 $. Now, we use **Hensel's Lemma** to lift this solution to modulo $ 289 $.

---

### Step 4: Lifting the Solution to Modulo 289

We seek an integer $ n $ such that:
$$
n \equiv 8 \mod 17 \quad \text{and} \quad n^4 \equiv -1 \mod 289
$$

Let $ n = 8 + 17k $. Then:
$$
n^4 = (8 + 17k)^4 = 8^4 + 4 \cdot 8^3 \cdot 17k + \text{(higher powers of 17)}
$$
Since higher powers of 17 are divisible by $ 289 $, we only need to consider the first two terms modulo 289.

So:
$$
n^4 \equiv 8^4 + 4 \cdot 8^3 \cdot 17k \mod 289
$$
Compute:
- $ 8^4 = 4096 \equiv 50 \mod 289 $
- $ 8^3 = 512 \equiv 223 \mod 289 $
- So:
$$
n^4 \equiv 50 + 4 \cdot 223 \cdot 17k \mod 289
$$
$$
\Rightarrow n^4 \equiv 50 + 136 \cdot 17k \mod 289
$$
$$
\Rightarrow n^4 \equiv 50 + 136 \cdot 17k \mod 289
$$

We want this to be $ \equiv -1 \mod 289 $, so:
$$
50 + 136 \cdot 17k \equiv -1 \mod 289
\Rightarrow 136 \cdot 17k \equiv -51 \mod 289
$$

Simplify:
- $ 136 \cdot 17 = 2312 \equiv 136 \mod 289 $
- So:
$$
136k \equiv -51 \mod 289
\Rightarrow 136k \equiv 238 \mod 289
$$

Now solve:
$$
136k \equiv 238 \mod 289
$$

We compute the inverse of 136 modulo 289. Since $ \gcd(136, 289) = 17 $, and 17 divides 238, we can divide the equation by 17:
$$
8k \equiv 14 \mod 17
$$

Find the inverse of 8 modulo 17. Try $ x = 15 $: $ 8 \cdot 15 = 120 \equiv 1 \mod 17 $, so inverse is 15.

So:
$$
k \equiv 14 \cdot 15 \mod 17 \equiv 210 \mod 17 \equiv 6 \mod 17
$$

Therefore:
$$
k = 6 + 17m
\Rightarrow n = 8 + 17k = 8 + 17 \cdot (6 + 17m) = 110 + 289m
$$

So the smallest positive solution is:
$$
n = 110
$$

---

### Step 5: Final Answer

We have shown that:

- The **smallest prime $ p $** for which there exists $ n $ such that $ n^4 + 1 $ is divisible by $ p^2 $ is $ p = 17 $.
- The **smallest positive integer $ m $** such that $ m^4 + 1 $ is divisible by $ 17^2 = 289 $ is $ m = 110 $.

---

### Final Answer

$$
\boxed{110}
$$
\end{Verbatim}
\end{aimebox}

%% file: Sections/Appendix_Sections/Samples/qwen_niah.tex
\begin{niahbox}{NIAH Multi-Key, 32K Context --- Qwen3-4B GDN Hybrid}
\samplelabel{Prompt}
\begin{Verbatim}[breaklines=true, breakanywhere=true, fontsize=\footnotesize, breaksymbolleft={}]
Some special magic numbers are hidden within the following text. Make sure to memorize them. I will quiz you about the numbers afterwards.

[... 14,350 filler sentences: "The grass is green. The sky is blu..." ...]
One of the special magic numbers for spirited-fiddle is: 8350860.
[... 2,075 filler sentences: "The grass is green. The sky is blu..." ...]
One of the special magic numbers for gentle-pillar is: 6351498.
[... 9,050 filler sentences: "The grass is green. The sky is blu..." ...]
One of the special magic numbers for noble-meadow is: 8411372.
[... 100 filler sentences: "The grass is green. The sky is blu..." ...]
One of the special magic numbers for luminous-ember is: 1144632.
[... 6,800 filler sentences: "The grass is green. The sky is blu..." ...]

What is the special magic number for luminous-ember mentioned in the provided text?
\end{Verbatim}
\samplelabel{Model output}
\begin{Verbatim}[breaklines=true, breakanywhere=true, fontsize=\footnotesize, breaksymbolleft={}]
The special magic number for **luminous-ember** mentioned in the text is **1144632**.
\end{Verbatim}
\end{niahbox}

%% file: Sections/Appendix_Sections/Samples/mimo_aime.tex
\begin{aimebox}{AIME 2024 I, Problem 14 --- MiMo-7B-RL-0530 GDN Hybrid}
\samplelabel{Problem}
Let $ABCD$ be a tetrahedron such that $AB=CD= \sqrt{41}$, $AC=BD= \sqrt{80}$, and $BC=AD= \sqrt{89}$. There exists a point $I$ inside the tetrahedron such that the distances from $I$ to each of the faces of the tetrahedron are all equal. This distance can be written in the form $\frac{m \sqrt n}{p}$, where $m$, $n$, and $p$ are positive integers, $m$ and $p$ are relatively prime, and $n$ is not divisible by the square of any prime. Find $m+n+p$.\hfill\textit{Answer: 104}
\samplelabel{Model reasoning (thinking mode)}
\begin{Verbatim}[breaklines=true, breakanywhere=true, fontsize=\footnotesize, breaksymbolleft={}]
Okay, so I have this problem about a tetrahedron ABCD with specific edge lengths, and I need to find the distance from an interior point I to each face, which is equal. The answer should be in the form m*sqrt(n)/p, and then find m + n + p. Hmm, let's start by understanding the problem step by step.

First, the tetrahedron has edges AB = CD = sqrt(41), AC = BD = sqrt(80), and BC = AD = sqrt(89). The point I is inside the tetrahedron and is equidistant from all four faces. That distance is called the inradius, right? Wait, but in a tetrahedron, the inradius is the radius of the sphere tangent to all four faces. But the problem says there exists such a point I, so maybe this tetrahedron is isohedral? Or maybe it's a special tetrahedron where such a point exists. Maybe it's a tetrahedron that's symmetric enough to have an inradius.

But how do I calculate the inradius? For a tetrahedron, the formula for the inradius r is given by r = 3V / S, where V is the volume of the tetrahedron and S is the total surface area. Wait, is that right? Let me recall. For a tetrahedron, the volume can be calculated using the scalar triple product, but maybe there's a more straightforward way here given the edge lengths.

[... 8,692 tokens of reasoning omitted ...]

Wait, vectors BC = C - B = (4-4, 0-5, 8-0) = (0, -5, 8)

Vectors BD = D - B = (0-4, 5-5, 8-0) = (-4, 0, 8)

Cross product BC x BD is determinant:

|i  j  k|
|0 -5  8|
|-4 0  8|

= i*(-5*8 - 8*0) - j*(0*8 - 8*(-4)) + k*(0*0 - (-5)*(-4))

= i*(-40 - 0) - j*(0 + 32) + k*(0 - 20)

= (-40, -32, -20)

Same as before. So magnitude is sqrt(1600 + 1024 + 400) = sqrt(3024) = 12*sqrt(21). Area is half that, so 6*sqrt(21). Correct.

So, all faces are indeed 6*sqrt(21), total surface area 24*sqrt(21). Volume 160/3. So, r = 3*(160/3)/(24*sqrt(21)) = 160/(24*sqrt(21)) = 20/(3*sqrt(21)) = 20*sqrt(21)/63. Correct.

Therefore, the answer is 20 + 21 + 63 = 104.

**Final Answer**
\boxed{104}
\end{Verbatim}
\samplelabel{Model answer}
\begin{Verbatim}[breaklines=true, breakanywhere=true, fontsize=\footnotesize, breaksymbolleft={}]
To solve the problem, we start by noting that the tetrahedron \(ABCD\) has edge lengths \(AB = CD = \sqrt{41}\), \(AC = BD = \sqrt{80}\), and \(BC = AD = \sqrt{89}\). We need to find the inradius of this tetrahedron, which is the distance from the interior point \(I\) to each face, given in the form \(\frac{m \sqrt{n}}{p}\).

First, we recognize that the tetrahedron can be embedded in a rectangular box with dimensions \(4 \times 5 \times 8\). This is confirmed by solving the system of equations derived from the edge lengths:
\[
x^2 + y^2 = 41, \quad x^2 + z^2 = 80, \quad y^2 + z^2 = 89.
\]
Solving these, we find \(x = 4\), \(y = 5\), and \(z = 8\).

Next, we calculate the volume \(V\) of the tetrahedron using the scalar triple product. The vectors from point \(A(0,0,0)\) to \(B(4,5,0)\), \(C(4,0,8)\), and \(D(0,5,8)\) are \(\overrightarrow{AB} = (4,5,0)\), \(\overrightarrow{AC} = (4,0,8)\), and \(\overrightarrow{AD} = (0,5,8)\). The cross product \(\overrightarrow{AB} \times \overrightarrow{AC}\) is:
\[
\overrightarrow{AB} \times \overrightarrow{AC} = \begin{vmatrix}
\mathbf{i} & \mathbf{j} & \mathbf{k} \\
4 & 5 & 0 \\
4 & 0 & 8
\end{vmatrix} = (40, -32, -20).
\]
The dot product with \(\overrightarrow{AD}\) is:
\[
(40, -32, -20) \cdot (0, 5, 8) = 0 - 160 - 160 = -320.
\]
The magnitude of this vector is \(\sqrt{40^2 + (-32)^2 + (-20)^2} = \sqrt{1600 + 1024 + 400} = \sqrt{3024} = 12\sqrt{21}\). The volume is:
\[
V = \frac{1}{6} \left| -320 \right| = \frac{320}{6} = \frac{160}{3}.
\]

The total surface area \(S\) is calculated by finding the area of each face. Each face is a triangle with area \(\frac{1}{2}\) the magnitude of the cross product of two edges. For example, for face \(ABC\), the vectors are \(\overrightarrow{AB} = (4,5,0)\) and \(\overrightarrow{AC} = (4,0,8)\), and the cross product is \((40, -32, -20)\). The magnitude is \(\sqrt{40^2 + (-32)^2 + (-20)^2} = 12\sqrt{21}\), so the area is \(\frac{1}{2} \times 12\sqrt{21} = 6\sqrt{21}\). Since all four faces are congruent, the total surface area is:
\[
S = 4 \times 6\sqrt{21} = 24\sqrt{21}.
\]

The inradius \(r\) is given by the formula \(r = \frac{3V}{S}\). Substituting the values, we get:
\[
r = \frac{3 \times \frac{160}{3}}{24\sqrt{21}} = \frac{160}{24\sqrt{21}} = \frac{20}{3\sqrt{21}} = \frac{20\sqrt{21}}{63}.
\]

Thus, the inradius is \(\frac{20\sqrt{21}}{63}\). Here, \(m = 20\), \(n = 21\), and \(p = 63\). These values are coprime and \(n\) is square-free. Therefore, \(m + n + p = 20 + 21 + 63 = 104\).

The final answer is:
\[
\boxed{104}
\]
\end{Verbatim}
\end{aimebox}

%% file: Sections/Appendix_Sections/Samples/mimo_niah.tex
\begin{niahbox}{NIAH Multi-Query, 32K Context --- MiMo-7B-RL-0530 GDN Hybrid}
\samplelabel{Prompt}
\begin{Verbatim}[breaklines=true, breakanywhere=true, fontsize=\footnotesize, breaksymbolleft={}]
Some special magic numbers are hidden within the following text. Make sure to memorize them. I will quiz you about the numbers afterwards.

[... 12,525 filler sentences: "The grass is green. The sky is blu..." ...]
One of the special magic numbers for vivid-walnut is: 8616278.
[... 4,900 filler sentences: "The grass is green. The sky is blu..." ...]
One of the special magic numbers for placid-lantern is: 7006521.
[... 1,375 filler sentences: "The grass is green. The sky is blu..." ...]
One of the special magic numbers for mellow-bramble is: 8084209.
[... 4,600 filler sentences: "The grass is green. The sky is blu..." ...]
One of the special magic numbers for rugged-ember is: 5916232.
[... 8,950 filler sentences: "The grass is green. The sky is blu..." ...]

What are the special magic numbers for vivid-walnut, placid-lantern, mellow-bramble, and rugged-ember mentioned in the provided text? State each key with its number.
\end{Verbatim}
\samplelabel{Model output}
\begin{Verbatim}[breaklines=true, breakanywhere=true, fontsize=\footnotesize, breaksymbolleft={}]
The special magic numbers mentioned in the text are as follows:

- **vivid-walnut**: 8616278  
- **placid-lantern**: 7006521  
- **mellow-bramble**: 8084209  
- **rugged-ember**: 5916232  

Each key is paired with its corresponding number as stated in the text.
\end{Verbatim}
\end{niahbox}